\documentclass[sigconf, 9pt]{acmart}
\pdfoutput=1

\AtBeginDocument{%
  \providecommand\BibTeX{{%
    \normalfont B\kern-0.5em{\scshape i\kern-0.25em b}\kern-0.8em\TeX}}}

\setcopyright{none} 

\acmConference[Submitted version to IPSN]{ACM/IEEE International Conference on Information Processing in Sensor Networks}{2023}{San Antonio, USA}
\acmPrice{xx.00}

\acmSubmissionID{\# 21}

\usepackage[utf8]{inputenc}
\usepackage[binary-units]{siunitx}
\usepackage{balance}
\usepackage{nicefrac} 
\usepackage{acronym}
\usepackage{xspace}

\usepackage{caption}
\usepackage{subcaption}
\newcommand{\aggregate}[2]{\underset{#2}{\operatornamewithlimits{#1\ }}}

\usepackage{multirow}
\usepackage{booktabs}

\usepackage{stmaryrd} 

\newcommand{\fakeparagraph}[1]{\vspace{.5mm}\textbf{#1.}}

\newcommand{\fakepar}[1]{\fakeparagraph{#1}}
\newcommand{\capt}[1]{\mdseries{\emph{#1}}}

\newcommand{\system}{DeepGANTT\xspace}

\newcommand{\dpr}[1]{}
\newcommand{\cp}[1]{}
\newcommand{\nt}[1]{}
\newcommand{\mb}[1]{}
\newcommand{\dk}[1]{}
\newcommand{\tv}[1]{}

\acrodef{TSCH}{Time-Slotted Channel Hopping}
\acrodef{RF}{Radio Frequency}
\acrodefindefinite{RF}{an}{a}
\acrodef{LNA}{Low-Noise Amplifier}
\acrodefindefinite{LNA}{an}{a}
\acrodef{LO}{Local Oscillator}
\acrodefindefinite{LO}{an}{a}
\acrodef{ADC}{Analog-to-Digital Converter}
\acrodef{IF}{Intermediate Frequency}
\acrodef{CMOS}{Complementary Metal-Oxide-Semiconductor}
\acrodef{DBP}{Digital Baseband Processor}
\acrodef{DSSS}{Direct Sequence Spread Spectrum}
\acrodef{ASK}{Amplitude Shift Keying}
\acrodef{MSK}{Minimum Shift Keying}
\acrodefindefinite{MSK}{an}{a}
\acrodef{FSK}{Frequency Shift Keying}
\acrodefindefinite{FSK}{an}{a}
\acrodef{oqpsk}[O-QPSK]{Offset-Quadrature Phase Shift Keying}
\acrodef{BPF}{Band-Pass Filter}
\acrodef{PRR}{Packet Reception Ratio}
\acrodef{SDR}{Software Defined Radio}
\acrodefindefinite{SDR}{an}{a}
\acrodef{IC}{Integrated Circuit}
\acrodef{WSN}{Wireless Sensor Networks}
\acrodef{COP}{Combinatorial Optimization Problem}
\acrodef{CCA}{Clear Channel Assessment}
\acrodef{MAC}{Medium Access Control}
\acrodef{IoT}{Internet of Things}
\acrodef{COTS}{Commercial Off-The-Shelf}

\acrodef{COP}{Combinatorial Optimization Problem}
\acrodef{ML}{Machine Learning}
\acrodefindefinite{ML}{an}{a}
\acrodef{DL}{Deep Learning}
\acrodef{GRL}{Graph Representation Learning}
\acrodef{GNN}{Graph Neural Network}
\acrodef{BN}{Bayesian Network}
\acrodef{RNN}{Recurrent Neural Network}
\acrodefindefinite{RNN}{an}{a}
\acrodef{NMT}{Neural Machine Translation}
\acrodefindefinite{NMT}{an}{a}
\acrodef{RL}{Reinforcement Learning}
\acrodefindefinite{RL}{an}{a}
\acrodef{DRL}{Deep Reinforcement Learning}
\acrodef{SSL}{Self-Supervised Learning}
\acrodefindefinite{SSL}{an}{a}
\acrodef{seq2seq}{Sequence-to-Sequence}
\acrodef{LSTM}{Long Short-Term Memory}
\acrodefindefinite{LSTM}{an}{a}

\begin{document}

\title{\system: A Scalable Deep Learning Scheduler for Backscatter Networks}


\author{Daniel F. Perez-Ramirez}
\orcid{0000-0002-1322-4367}
\affiliation{%
  \institution{RISE Computer Science \& \\ KTH Royal Institute of Technology}
  \city{Stockholm}
  \country{Sweden}
}
\email{daniel.perez@ri.se}

\author{Carlos Pérez-Penichet}
\orcid{0000-0002-1903-4679}
\affiliation{%
  \institution{RISE Computer Science}
  \city{Stockholm}
  \country{Sweden}
}
\email{carlos.penichet@ri.se}

\author{Nicolas Tsiftes}
\orcid{0000-0003-3139-2564}
\affiliation{%
  \institution{RISE Computer Science}
  \city{Stockholm}
  \country{Sweden}
}
\email{nicolas.tsiftes@ri.se}

\author{Thiemo Voigt}
\orcid{0000-0002-2586-8573}
\affiliation{%
  \institution{Uppsala University \& \\ RISE Computer Science}
  \city{Uppsala \& Stockholm}
  \country{Sweden}
}
\email{thiemo.voigt@angstrom.uu.se}

\author{Dejan Kostić}
\orcid{0000-0002-1256-1070}
\affiliation{%
  \institution{KTH Royal Institute of Technology \& RISE Computer Science} 
  \city{Stockholm}
  \country{Sweden}
}
\email{dmk@kth.se}

\author{Magnus Boman}
\orcid{0000-0001-7949-1815}
\affiliation{%
  \institution{KTH Royal Institute of Technology}
  \city{Stockholm}
  \country{Sweden}
}
\email{mab@kth.se}

\renewcommand{\shortauthors}{Perez-Ramirez, et al.}

\renewcommand{\shortauthors}{Perez-Ramirez, et al.}

\begin{abstract}
    Novel backscatter communication techniques enable battery-free sensor tags to interoperate with unmodified standard IoT devices, extending a sensor network's capabilities in a scalable manner. 
    Without requiring additional dedicated infrastructure, the battery-free tags harvest energy from the environment, while the IoT devices provide them with the unmodulated carrier they need to communicate.
    A schedule coordinates the provision of carriers for the communications of battery-free devices with IoT nodes.
    Optimal carrier scheduling is an NP-hard problem that
    limits the scalability of network deployments. Thus, existing solutions waste energy and other valuable resources by scheduling the carriers suboptimally.
    We present \system, a deep learning scheduler that leverages graph neural networks to efficiently provide near-optimal carrier scheduling. 
    We train our scheduler with relatively small optimal schedules obtained from a constraint optimization solver, achieving a performance within 3\% of the optimal scheduler.
    \system generalizes to networks $6\times$ larger in the number of nodes and $10\times$ larger in the number of tags than those used for training, breaking the scalability limitations of the optimal scheduler and reducing carrier utilization by up to $50\%$ compared to the state-of-the-art heuristic. Our scheduler efficiently reduces energy and spectrum utilization in backscatter networks.
\end{abstract}



\settopmatter{printfolios=true}
\settopmatter{printacmref=false}

\maketitle

\begin{figure}[!t]
    \centering
    \includegraphics[width=\linewidth]{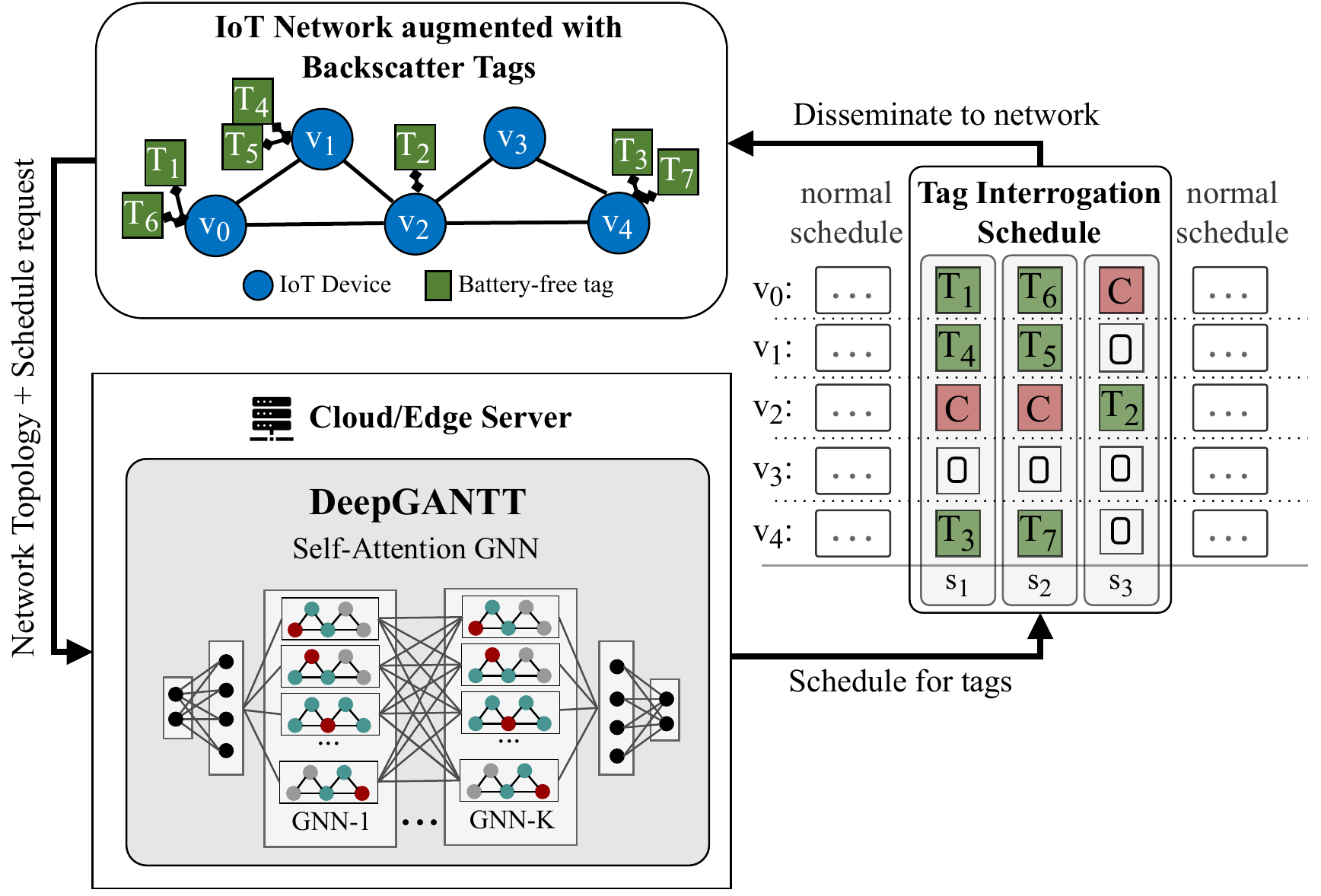}
    \caption{\capt{\system employs \acs*{GNN}s to
      schedule wireless communications in a backscatter network.} 
      It takes a graph representing the wireless network as input and produces a schedule, which directs IoT nodes ($v$) to interrogate every battery-free tag ($\mathtt{T}$) with minimal resources by reducing the number of carriers ($\mathtt{C}$) and timeslots ($s$) needed. 
      }
    \label{fig:system-overview}
    \vspace*{-0.3cm}
\end{figure}

\begin{figure}[!t]
    \centering
    \includegraphics[width=0.96\linewidth]{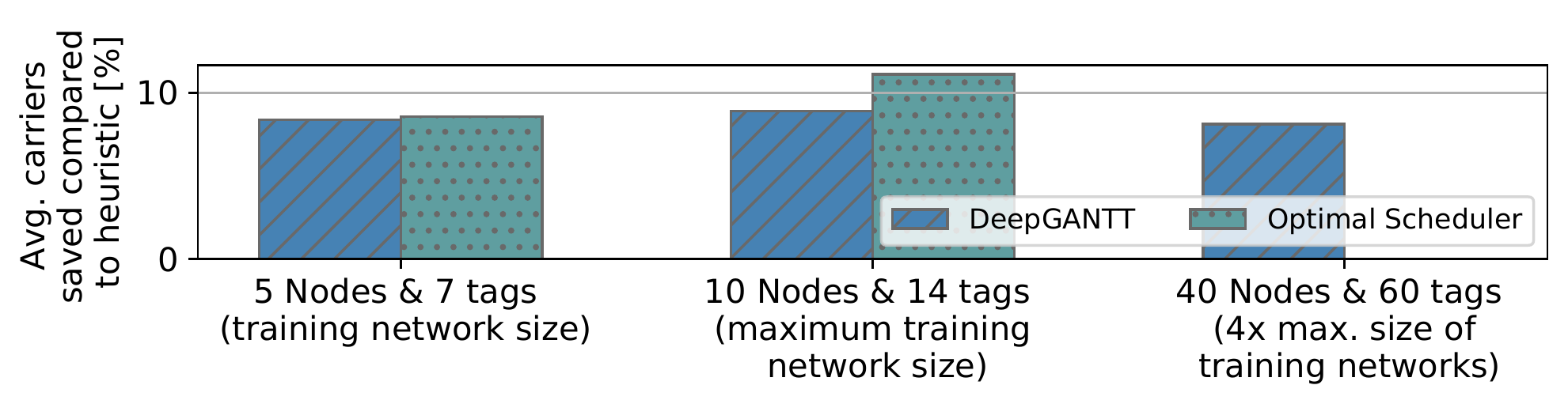}
    \caption{\capt{\system provides near-optimal schedules and
    generalizes well beyond the scalability limitation of the optimal
    scheduler.}  \system's performance is within 3\% of the optimum
    on trained problem sizes, but can be applied to much larger
    networks.} 
    \label{fig:carrs-saved}
    \vspace*{-0.4cm}
\end{figure}

\section{Introduction}
Backscatter communications enable a new class of wireless devices that harvest energy from their environment to operate without batteries~\cite{Liu2013ambient,fmbackscatter,hessar_netscatter_2018}. 
Recent advances have demonstrated how these battery-free backscatter devices---\emph{tags} for short--- can seamlessly
perform bidirectional communications with unmodified \ac{COTS} wireless
devices over standard physical layer protocols when assisted by an external unmodulated carrier
~\cite{kellogg_passive_2016,ensworth_every_2015,kellogg2014wi,talla2017lora,iyer_inter-technology_2016,Perez-Penichet2016augmenting,perez-penichet_tagalong_2020}.
This synergy facilitates new applications where tags are placed in hard-to-reach locations to perform sensing without the encumbrance of bulky
batteries and the maintenance cost of frequent battery
replacements~\cite{Perez-Penichet2016augmenting,perez-penichet_tagalong_2020}.
However, for backscatter tags to communicate, the \ac{IoT} devices in the network must cooperate to provide
an unmodulated \ac{RF} carrier. Unfortunately, providing carrier support means a significant resource investment for the \ac{IoT} devices, that may be battery-powered. As a consequence, the efficient provision of unmodulated carriers is of paramount importance for system-level performance and network lifetime.
%

\fakepar{Scenario} We consider the scenario where a network of \ac{COTS} wireless \ac{IoT} devices
has been augmented with battery-free tags that do sensing on their
behalf~\cite{Perez-Penichet2016augmenting}. In this context, one can see the tags as devices that wirelessly provide additional functionality to the \ac{IoT} nodes with simplicity akin to adding Bluetooth peripherals to our computers, but without incurring extensive maintenance and deployment costs associated with battery-powered nodes~\cite{Perez-Penichet2016augmenting,perez-penichet_tagalong_2020}. An advantage of this \emph{tag-augmented} 
scenario is that the battery-free tags can be located in hard-to-reach locations such as moving machinery, medical implants, or embedded in walls and floors.
Meanwhile, the more capable \ac{IoT} devices are placed in accessible locations
nearby, where either battery replacement or mains power is
available~\cite{Perez-Penichet2016augmenting,perez-penichet_tagalong_2020}. The
\ac{IoT} nodes perform their own sensing, communication, and computation according to their normal schedule. Additionally, a \emph{tag interrogation schedule} is required for the commodity devices to collect sensor readings from the tags by coordinating among themselves to provide the unmodulated carrier that tags need to both receive and transmit data (see Figure~\ref{fig:system-overview}).  

Consider, for instance, a healthcare monitoring application that includes
implanted and wearable sensors~\cite{jameel2019applications}. If the
battery-powered wearables cooperate to collect measurements from the implants,
they could spare the patients from undergoing surgery just to replace the
implants' batteries because they can now be battery-free. 
Making these devices battery-free is also important for sustainability
reasons. Similar examples can be envisioned for applications such as industrial machinery, smart agriculture, and infrastructure monitoring~\cite{Varshney2017lorea, Hester2018futureOfSensing}.

\fakepar{Challenges} 
\begin{figure}
    \centering
     \includegraphics[width=0.9\linewidth]{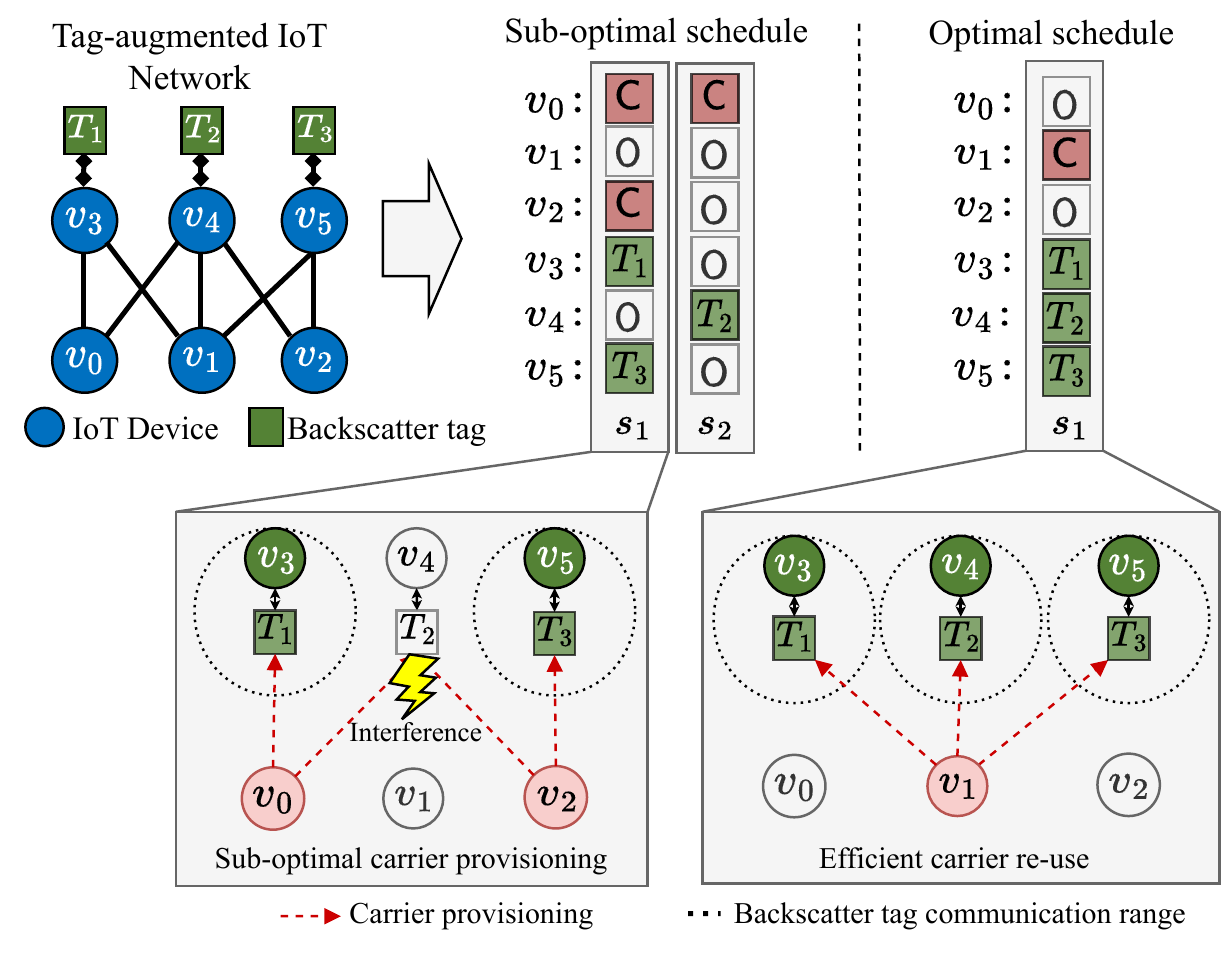}
     \caption{\capt{Optimal schedules favor maximum carrier re-use}.
        Example of a tag-augmented IoT network and two possible schedules: the optimal schedule that maximizes carrier re-use while minimizing schedule length, and a sub-optimal schedule.
        }
     \label{fig:carrier-reuse}
     \vspace*{-0.5cm}
\end{figure}
In tag-augmented scenarios, \ac{IoT} devices invest considerable resources to support the tags, despite the fact that they may be battery-powered or otherwise resource-constrained. 
%
To minimize the resources allocated to supporting the tags, one must devise an efficient communication schedule for tag interrogations~\cite{perez-penichet_tagalong_2020,PerezPenichet2020afast}.
A single tag interrogation 
corresponds to a request-response cycle between an \ac{IoT} device and one of its hosted battery-free sensor tags,
while exactly one of its neighboring \ac{IoT} nodes provides an unmodulated carrier. More than one unmodulated
carrier impinging on any tag causes interference and prevents proper interrogation. 
Provided that collisions are avoided, scheduling multiple tag interrogations
concurrently reduces the schedule's length, improving latency
and throughput in the network (see Figure~\ref{fig:carrier-reuse}). Furthermore, scheduling one unmodulated
carrier to serve multiple tags simultaneously greatly reduces the required energy and spectrum occupancy. 

Computing the optimal schedule---minimizing the number of times nodes must
generate carriers and minimizing the duration of the schedule---is an
NP-hard combinatorial optimization problem~\cite{PerezPenichet2020afast}.
At first glance, tag scheduling is similar to classic wireless link scheduling
in that collisions among data transmissions must be avoided by
considering the network topology. Carrier scheduling differs, however, in that we must additionally select appropriate carrier generators
while avoiding collisions caused by them upon their neighbors. Simultaneously, we must also minimize resource utilization. 
The only known scalable solution is a carefully crafted heuristic, which nevertheless suffers from suboptimal performance~\cite{PerezPenichet2020afast}. This results in wasted energy and spectral resources, particularly as network sizes grow. 
We refer to this heuristic as the \emph{TagAlong scheduler}~\cite{perez-penichet_tagalong_2020} hereafter. Alternatively, one can compute optimal solutions using a constraint optimization solver. 
We will refer to this scheduler as the \emph{optimal scheduler}.
This scheduler, however, takes a prohibitively long time 
as network sizes grow, which limits the capacity to adapt to network topology
changes. E.g., computing schedules for 10 nodes and 14 tags can take up
to a few hours.

In this paper, we leverage \ac{DL} methods to overcome both the scalability
limitations of the optimal scheduler and the performance shortcomings of the
TagAlong scheduler. However, the \ac{DL} approach presents its own set of
challenges.
First, traditional \ac{DL} methods that have succeeded on problems with
fixed structure and input/output sizes (e.g., images and tabular
data) are not applicable to the carrier scheduling problem since the latter operates on an irregular network structure, the size of which depends on the number of \ac{IoT} devices and sensor tags.
Second, any given \ac{IoT} network configuration may have many equivalent optimal solutions, which can confuse \ac{ML} models during
training. This is due to symmetries inherent to the carrier scheduling problem, e.g., the schedules are invariant to timeslot permutations when not imposing a fix priority on tag interrogation.

\fakepar{Approach}
In this work, we present \emph{Deep Graph Attention-based Network Time Tables} (\system), a new scheduler that 
builds upon the most recent advances in \ac{DL} to 
efficiently schedule
the communications of battery-free tags and the supporting carrier generation in a heterogeneous network of \ac{IoT} devices interoperating with battery-free tags.
%
Upon request from the wireless network, \system receives as input the network topology represented as a graph, and generates a corresponding interrogation schedule~(Figure~\ref{fig:system-overview}). The graph representation of the network assumes there is a link between two IoT nodes iff there is a sufficient signal strength between them for providing unmodulated carrier. 

\system iteratively performs one-shot node classification of the network's \ac{IoT} devices to determine the role each of them will play within every schedule timeslot:
either remain
off ($\mathtt{O}$), interrogate one of its tags ($\mathtt{T}$), or generate a
carrier ($\mathtt{C}$), while also avoiding collisions in the network. 
The objective of the carrier scheduling problem is to reduce the resources needed to interrogate every tag in the network. By minimizing the number of carrier generation slots ($\mathtt{C}$) in the schedule, we reduce energy and spectrum occupancy. As a secondary objective, minimizing the number of required timeslots improves latency and throughput.

We adopt a supervised learning approach based on \acp{GNN} instead of other paradigms such as reinforcement learning.  This choice is mainly motivated by three facts. 
First, 
we can leverage the optimal scheduler to generate the training data 
necessary for a supervised approach.
Second, 
\acp{GNN} are particularly successful in handling irregularly structured,
variable-size input data such as network topology
graphs~\cite{Scarselli2009gnnmodel, Gilmer2017mpnn, Kipf2017gcn}. \acp{GNN} provide a natural way of capturing the interdependence among neighboring nodes across multiple hops, crucial to avoid collisions.
Third, 
it is straightforward to cast the scheduling problem as a classification
task, which is generally tackled with a supervised approach~\cite{bishop2006pattern, murphy2012machine}.

To train \system, 
we generate random network topologies
that are small enough for the optimal scheduler to handle. To avoid the pitfalls of training with multiple optimal solutions per instance, we add symmetry-breaking constraints to the optimal scheduler. Such constraints alter neither the true constraints, nor the objective of the problem. Instead, they narrow the choices of the solver from potentially many equivalent optimal solutions down to a single consistent one.
For example, in tag scheduling, the order in which tags are interrogated is irrelevant. Nevertheless, by adopting a specific order (e.g., decreasing order of tag ID) we reduce the number of solutions from the factorial of the number of tags to one.


\fakepar{Contributions} 
We make the following specific contributions:
\begin{itemize}
    \item We present \system, the first fast and scalable \ac{DL} scheduler that leverages \acp{GNN} to obtain near-optimal solutions to the carrier scheduling problem. 
        %
    \item We employ symmetry-breaking constraints to limit the
        solution space of the carrier scheduling problem when generating the training data.
    \item \system performs within $3\%$ of the optimum in trained network sizes and scales to $6\times$ larger networks while reducing carrier generation slots by up to $50\%$ compared to the state-of-the-art heuristic. This directly translates to energy and spectrum savings.
        %
    \item We use \system to compute schedules for a real network topology of \ac{IoT} devices. Compared to the heuristic, our scheduler reduces the energy per tag interrogation by $13\%$ in average and up to $50\%$ for large tag deployments.  
    \item \system's inference time is polynomial on the input size, achieving on average \SI{429}{\ms} and always below \SI{1.5}{\s}; a radical improvement over the optimal
        scheduler. 
\end{itemize}
Hence, we show that our scheduler can compute more resource-efficient
schedules than the TagAlong scheduler, and that these are
almost as good as those of the optimal scheduler (see Figure~\ref{fig:carrs-saved}). 
Moreover, \system can generate schedules for considerably
larger backscatter networks while still reducing the energy consumption and spectrum occupancy compared to the heuristic. Our scheduler also breaks the scalability limitations of the optimal scheduler, facilitating timely reactions to changes in topology and radio propagation conditions. 

%
The rest of the paper is organized as follows. Sec.~\ref{sec:background}
gives background information that is useful to understand the paper. Sec.~\ref{sec:probform} formally describes the carrier scheduling problem. Sec.~\ref{sec:sysdescrip} details the design of \system, while
Sec.~\ref{sec:learning-to-schedule} discusses the implementation and training of the model. In Sec.~\ref{sec:evaluation}, we evaluate \system's performance against previous alternatives and prove that it can also compute schedules for a real IoT network. Lastly, we discuss related work in Sec.~\ref{sec:related_work} and conclude our work in Sec.~\ref{sec:conclusion}.

\section{background}\label{sec:background}

\begin{figure}
    \centering
     \begin{subfigure}[b]{0.45\linewidth}
         \centering
         \includegraphics[width=0.71\linewidth]{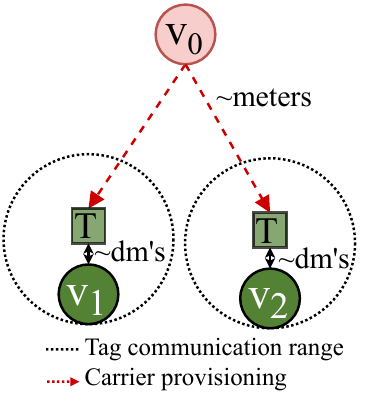}
         \caption{\capt{Carrier re-use enabled by tags' short communication range}. }
         \label{subfig:carrier-range}
     \end{subfigure}
     \hfill
     \begin{subfigure}[b]{0.50\linewidth}
         \centering
         \includegraphics[width=0.80\linewidth]{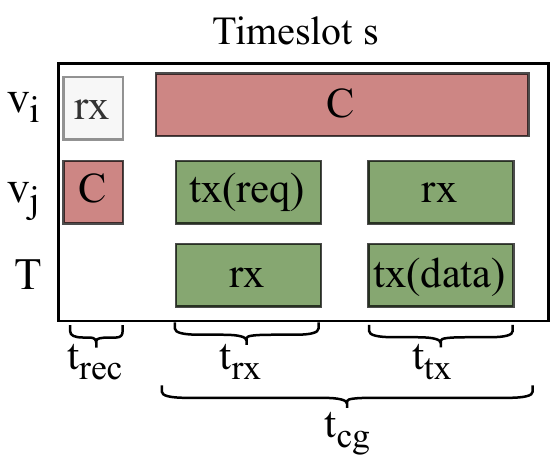}
         \caption{\capt{Tag uplink and downlink communication within a timeslot $s$}.}
         \label{subfig:timeslot-spec}
     \end{subfigure}
    \centering
    \caption{\capt{The duration of a timeslot is sufficient for an \ac{IoT} device to interrogate one tag by transmitting a request directed to the desired tag and receiving the response}. Furthermore, we leverage the short tag communication range to use a carrier providing node to perform multiple tag interrogations across nodes in the network.}
    \label{fig:timeslot-description}
    \vspace*{-0.5cm}
\end{figure}

This section gives a quick overview of backscatter communications and the
concept of \emph{tag-augmented} \ac{IoT} networks, with an intuitive
introduction to \ac{GNN}s.

\subsection{Backscatter Communications}\label{sec:backscatter_primer}
Backscatter communication devices are highly attractive because their
characteristic low power consumption enables them to operate without batteries.
Instead, they can sustain themselves by collecting energy from their environment
using energy harvesting modalities. 
These kinds of devices achieve their low power consumption by offloading some of
the most energy-intensive functions, such as the local oscillator, to an
external device that provides an unmodulated carrier. Recent works have extended
this principle to enable battery-free tags capable of direct two-way
communications with unmodified \ac{COTS} wireless devices using standard
protocols such as IEEE~802.15.4/ZigBee or Bluetooth, when supported by an
external 
carrier~\cite{kellogg_passive_2016,ensworth_every_2015,kellogg2014wi,talla2017lora,iyer_inter-technology_2016,Perez-Penichet2016augmenting,perez-penichet_tagalong_2020}.

Previous works have demonstrated systems that apply these battery-free
communication techniques to augment an existing network of \ac{COTS} wireless devices with battery-free
tags~\cite{Perez-Penichet2016augmenting,perez-penichet_tagalong_2020,PerezPenichet2020afast}.
We refer to this architecture as a \emph{tag-augmented} \ac{IoT} network. It enables placing
sensors in hard-to-reach locations without having to worry about wired energy
availability or battery maintenance.  Existing studies 
have shown how the
\ac{IoT} nodes invest energy to provide carrier support in proportion to the
number of carrier slots ($\mathtt{C}$) scheduled and that tags add latency
proportionally to the duration of the tag interrogation
schedule~\cite{PerezPenichet2020afast}. 
As a consequence, it is critical that we optimize the way unmodulated carrier
support is provided (see Figure~\ref{fig:carrier-reuse}). For this reason, in this work, we focus on the efficiency of the scheduler, given that it  bears total influence on resource expenditure.


We adopt a model where each tag is associated with (or hosted by) one
\ac{IoT} device responsible for interrogating it to collect sensor readings.  Every \ac{IoT}
node in the network may host zero or more tags. The \ac{IoT} devices in the
network are equipped with radio transceivers that support standard physical
layer protocols such as IEEE~802.15.4/ZigBee or Bluetooth. They are able to
provide an unmodulated carrier (by using their radio test
mode~\cite{Perez-Penichet2016augmenting}) and employ a time-slotted medium
access mechanism. The use of a time-slotted access mechanism is motivated by its widespread use in commodity devices and its ease of integration for the battery-free tags. 
The duration of a timeslot is sufficient for an \ac{IoT}
device to interrogate one tag by transmitting a request directed to the desired tag and
receiving the response. 
Figure~\ref{subfig:timeslot-spec} shows a tag interrogation procedure. First, the carrier providing node $\text{v}_i$ listens for a period $t_{req}$ for a request from the interrogating node $\text{v}_j$ at its assigned timeslot in the schedule. Upon the request arrival, $\textbf{v}_i$ provides a carrier for the duration $t_{cg}$. The interrogating node $\text{v}_j$ transmits its request to one of its hosted tags $T$, after which the tag transmits its response back to $\text{v}_j$. 

When a node interrogates a tag, one of its neighboring \ac{IoT} nodes must provide an unmodulated carrier~\cite{perez-penichet_tagalong_2020,PerezPenichet2020afast}. Note that both
IEEE~802.15.4 and Bluetooth specify a time-slotted access mechanism with the
necessary characteristics in their respective
standards~\cite{ieee802_15_4_2016,bluetooth_2021}. As shown in Figure~\ref{subfig:carrier-range}, the short communication range of the tags enables re-using an unmodulated carrier to perform multiple concurrent tag interrogations. However, an \ac{IoT} node can only interrogate one of its hosted tags per timeslot. When not querying the backscatter tags, the network of \ac{IoT} nodes performs its own tasks according to its regular schedule.
Finally, at least one of the \ac{IoT} devices is connected to a cloud or edge
server where the interrogation schedule can be computed.

The network of \ac{IoT} nodes keeps track of link state information 
to determine the connectivity graph among themselves. This information can be
relayed periodically or on demand
to the cloud or edge server, where it is, together with the tag-to-host mapping, 
used to assemble the graph representation of the network. 
Our scheduler uses this graph representation to produce a schedule, 
as depicted in
Figure~\ref{fig:system-overview}.  An interrogation schedule consists of one or
more scheduling \emph{timeslots}, each assigning one of three roles to every
\ac{IoT} device in the network: provide an
unmodulated carrier ($\mathtt{C}$), interrogate one of its hosted sensor tags ($\mathtt{T}$), or
remain idle ($\mathtt{O}$).

\subsection{Graph Neural Networks} \label{sec:background-gnn}

\acp{GNN} have emerged as a flexible means to tackle
various inference tasks on graphs, such as node
classification~\cite{Scarselli2009gnnmodel, hamilton2020graph, zonghan2021gnnsurvey}. 
Intuitively, 
stacking $K$ \ac{GNN} layers corresponds to generating node embedding vectors
taking into account
its $K$-hop neighborhood by leveraging the structure
of the graph and inter-node dependencies~\cite{Gilmer2017mpnn, Kipf2017gcn}.
These embeddings are typically further processed with linear
layers to produce the final output according to the task of interest. E.g., one might perform node classification by passing each node embedding vector through a classification layer.
Formally, given a graph $G=\langle V,E \rangle$ defined by the sets of nodes
$v\in V$ and edges $(v, u)\in E$, at \ac{GNN} layer $i$ each node feature
vector $h_v$ is updated
as: \begin{equation}
    h^{(i)}_v = f_1\left(\,\,h^{(i-1)}_v,\,\, \aggregate{AGG}{u\in \mathcal{N}(v)}\left[ f_2\left(h_u^{(i-1)}\right) \right] \,\,\right) \, \text{,}
    \label{eq:message-pas}
\end{equation}
where $\mathcal{N}(v)$ is the set of neighboring node feature vectors of
node $v$, AGG is a commutative aggregation function, and
$f_{1}, f_{2}$ are non-linear transformations~\cite{Gilmer2017mpnn}.
Among the main advantages of using \ac{GNN}s over traditional \ac{DL} methods
are their capability to exploit the structural dependencies of the graph. Furthermore, they are an inductive reasoning method, i.e., \acp{GNN}  can be deployed to perform inference on graphs other than those seen during training without the need to re-train the model~\cite{Hamilton2017inductive, Velickovic2018gat, vesselinova2020learning}.

\section{Problem Formulation}\label{sec:probform}
This section formally describes the carrier scheduling problem, i.e.,
efficiently scheduling the communications of sensor tags and the supporting carrier generation in a network of \ac{IoT} wireless devices interoperating with battery-free tags.
We hereon
refer to the \ac{IoT} devices and to the sensor tags in the network simply as
\emph{nodes} and \emph{tags}, respectively.
We model the wireless \ac{IoT} network as an undirected connected graph $G$, defined by the tuple $G=\langle V_a,E \rangle$, where $V_a$ is the set of $N$ nodes in the network $V_a=\{v_i\}_{i=0}^{N-1}$, and $E$ is the set of edges between the nodes $E=\{\langle u,v\rangle|u,v \in V_a\}$. 

The connectivity among nodes in the graph (edges set $E$) is determined by the
link state information collected as described in Section~\ref{sec:backscatter_primer}, i.e., there
is an edge between two nodes if and only if there is a sufficiently strong
wireless signal for providing the unmodulated
carrier~\cite{perez-penichet_tagalong_2020, PerezPenichet2020afast}.
We denote the set of $T$ tags in the network as $N_t=\{t_i\}_{i=0}^{T-1}$, and their respective tag-to-host assignment as $H_t: t\in N_t\mapsto v\in V_a$. 
A node can host zero or more tags.
The role of a node $v$ within a timeslot $s$ 
is indicated by the map $R_{v,s}: v\!\in \!V_a, s\!\in \![1, L]
\mapsto \{\mathtt{C}, \mathtt{T}, \mathtt{O}\}$, where $L$ is the schedule
length in timeslots. Hence, a timeslot $s_j$ consists of an $N$-dimensional
vector containing the roles assigned to every node during timeslot $j$:
$s_j=\left[R_{v_i,j} | v_i\in V_a \right]^{\top}$.
A timeslot duration is long enough to complete one interrogation
request-response cycle between a node and a tag (see Figure~\ref{subfig:timeslot-spec}).

For a given problem instance (wireless network configuration) defined by the
tuple $g=\langle G, N_t, H_t\rangle$,
the \ac{COP} of interrogating all sensor values in the
network once using the lowest number of carrier generators and timeslots is
formulated as follows:
\setlength{\arraycolsep}{0.0em}
\begin{eqnarray}
\min\,\,&{}&\,\,\left(T \times C + L \right) \label{eq:optobj}\\
\text{s.t.}\,\,&{}& \,\,\forall t\!\in\!N_t \,\,\exists!\,\,s\!\in\![1, L]:\, R_{H_t,s}=\mathtt{T}\label{eq:optconst1}\\
\,\,&{}& \,\,\forall s\!\in\![1, L]\,\,\forall t\!\in\!N_t\,|\,R_{H_t,s}=\mathtt{T} \label{eq:optconst2}\\ 
\,\,&{}& \,\,\;\;\;\;\;\;\;\;\;\;\; \exists!\,v_j\!\in\!V_a:\, R_{v_j,s}=\mathtt{C} \wedge (H_t, v_j)\in E \,\,\text{,} \nonumber
\end{eqnarray}
\setlength{\arraycolsep}{5pt}
where $C$ is the total number of carriers required in the schedule, i.e.,
$C=\left| \{R_{v_j,s}=\mathtt{C}: v_j\!\in\!V_a, s\!\in\![1,
L]\}\right|$. 
Constraints (\ref{eq:optconst1})
and (\ref{eq:optconst2}) enforce that tags are interrogated only once in the
schedule and that there is only one carrier-providing neighbor per tag in each
timeslot (to prevent collisions), respectively.
The objective function (\ref{eq:optobj}) is designed to prioritize reducing the
number of carrier slots~($C$) over the duration of the schedule~($L$). This is
because we are most concerned with energy and spectrum efficiency and because a
reduction of $C$ often implies a reduction of $L$, but the converse is not
necessarily true.  For example, in Figure~\ref{fig:system-overview}, $v_2$
provides a carrier to interrogate $T_1$ and $T_3$, reducing $C$ and $L$
simultaneously.  However, another alternative would be to provide carriers to
$T_1$ and $T_3$ from $v_1$ and $v_2$ respectively; which would reduce $L$ but
not $C$.

\section{\system: System Design}\label{sec:sysdescrip}

In this section, we first introduce the design considerations for \system to cope with the wireless network requirements and the challenges in carrier scheduling. We then introduce \system's system architecture.

\subsection{Design Considerations}
\system is deployed at an edge or cloud server, where schedules are computed on demand for the \ac{IoT} network. At least one of the \ac{IoT} nodes in the network is assumed to be connected to the Edge/Cloud server, and it is responsible for building the \ac{IoT} network topology graph, emitting the request to the scheduler in the Edge/Cloud, and disseminate the computed schedule to the other devices. 
The \system scheduler receives as input 
the \acp{IoT} network configuration as the tuple $g=\langle G,
N_t, H_t\rangle$, i.e., the wireless network topology $G$ 
and the set of tags $N_t$ in the network with their respective tag-to-host assignment $H_t$.
The scheduler then generates the interrogation schedule $S$ and delivers it
to the \ac{IoT} network. The scheduler may receive subsequent schedule requests by the \ac{IoT} network
either upon addition/removal of nodes or tags, or upon connectivity changes
among the \ac{IoT} devices. Thus, \system must be able to react fast to
structural and connectivity changes in the wireless network.

From \iac{ML} perspective, every possible configuration of $g$ yields a
different graph representation with different connectivity and potentially
different input size. Likewise, the output (schedule $S$) may vary in size in terms of both the number of timeslots $L$ and the number of nodes $N$ in the topology (since $s_j\!\in\!\mathbb{R}^{N} \forall j\in\{0, \dots L\}$).
Moreover, at every timeslot $s_j$, every node is assigned to one of three
possible actions $\{\mathtt{C}, \mathtt{T}, \mathtt{O}\}$ (see
Section~\ref{sec:backscatter_primer}) based on its neighborhood.  We use
a \ac{GNN}-based learning approach to allow \system to process variable-sized
inputs (network topology) and output (interrogation schedule) sequences while
learning the local dependencies of a node in the network 
(see Sec.~\ref{sec:background-gnn}).

For carrier scheduling, exploiting the structural dependencies of the nodes' $K$-hop neighborhood allows to efficiently schedule carriers while avoiding interference. 
Moreover, since \acp{GNN} are an inductive reasoning \ac{ML} method, we can deploy the model without the need to re-train it for different network configurations in terms of the number of \ac{IoT} nodes $N$, their connectivity, and the number of sensor tags $T$. 
%

\subsection{System Architecture} \label{subsec:sysarch}

\begin{figure}
    \begin{subfigure}[b]{0.9\linewidth}
        \includegraphics[width=0.9\linewidth]{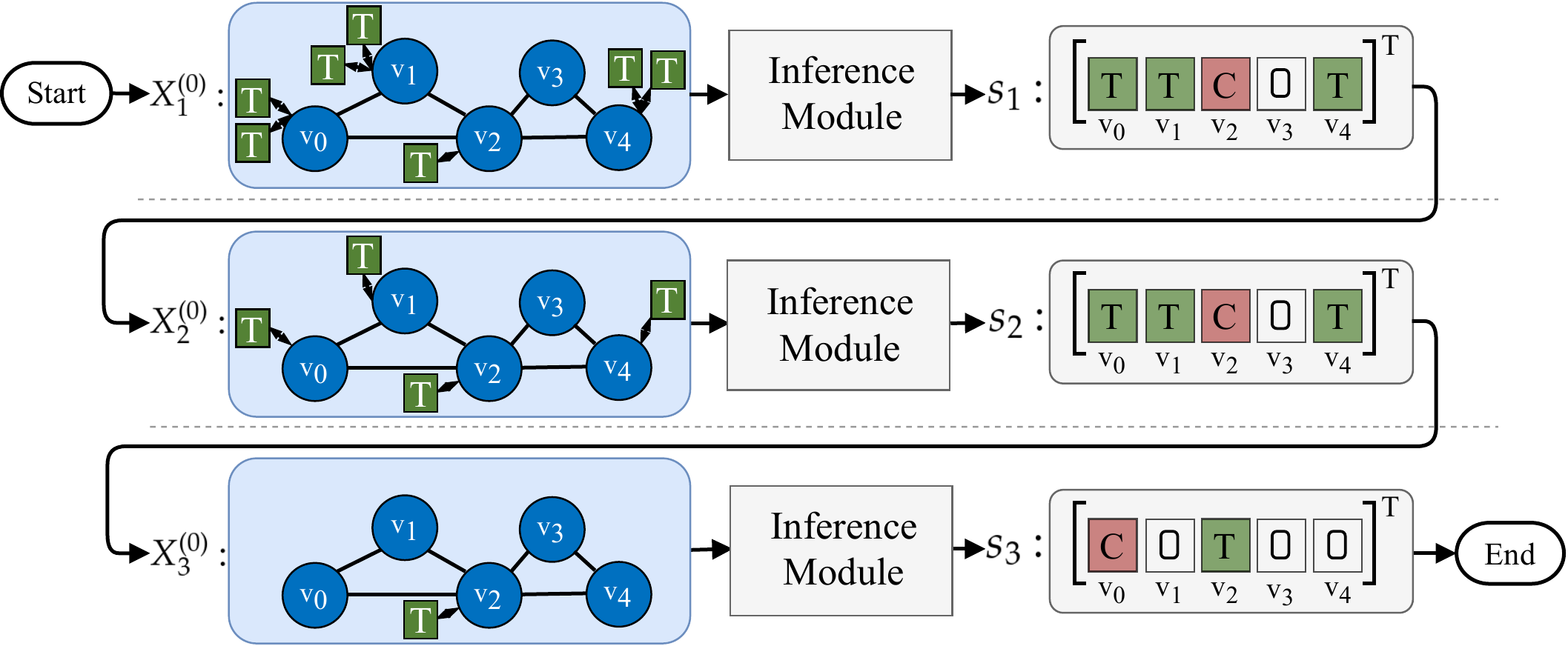}
        \caption{\capt{\system iteratively performs one-shot node classification, one timeslot at a time, removing scheduled tags from the topology and repeating the process until no more tags remain.}
        }
        \label{fig:pred-flow}
     \end{subfigure}
     \\[0.3cm]
     \centering
     \begin{subfigure}[b]{\linewidth}
         \centering
         \includegraphics[width=0.48\linewidth]{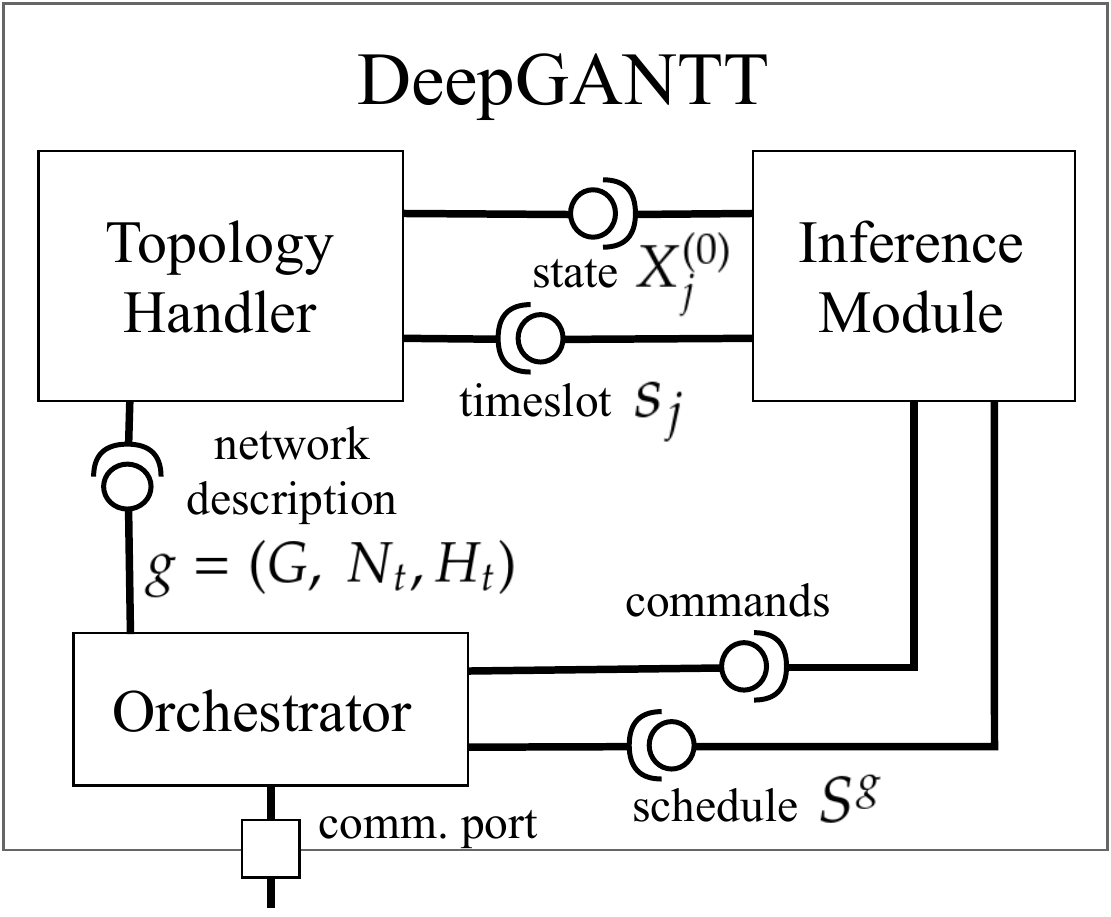}
         \caption{\capt{The three core components of \system interact among
         themselves and with the exterior to generate schedules.}}
         \label{subfig:sl-arch}
     \end{subfigure}
     \caption{\capt{\system's system architecture and the Inference Module's procedure to generate schedules.}}
    \label{fig:sl-sys}
    \vspace*{-0.5cm}
\end{figure}

We model the carrier scheduling problem as an iterative one-shot node classification problem. The inner workings of \system are
illustrated in Figure~\ref{fig:pred-flow}: on each iteration $j$, \system's
inference module assigns each node in the topology to one of three classes
corresponding
to the possible node actions: $\{\mathtt{C}, \mathtt{T}, \mathtt{O}\}$. Hence,
each iteration $j$ generates a timeslot $s_j$.
Each predicted timeslot is checked for compliance with the constraints in Eq.~\ref{eq:optconst1} and ~\ref{eq:optconst2}.
After each iteration,
the topology's 
node feature vector
is updated by removing one tag from the nodes
that were assigned class $\mathtt{T}$ (interrogate). This process is repeated until all tags have been removed from the cached network configuration. 

The \system scheduler consists of three submodules, as depicted in Figure~\ref{subfig:sl-arch}.
The \emph{Orchestrator} is \system's coordinating unit and interacts with
the outside world through its communication port. It is responsible for
providing the problem description to the \emph{Topology Handler} and interfacing
with the \emph{Inference Module}. 
The \emph{Topology Handler}
maintains the problem description, provides it to the \emph{Inference Module} in a format suitable for the \ac{ML} model, and updates its state 
according to the predicted
timeslots. 

Since the \emph{Inference Module} is trained
on the basis of a stochastic process, the \emph{Topology Handler} includes a
fail-safe functionality to make sure that the predictions comply with the
constraints in Eqs.~(\ref{eq:optconst1}) and~(\ref{eq:optconst2}). In case of
failure, the \emph{Topology Handler} restores compliance by randomly shuffling
tag and node IDs and retrying. This does not alter the final schedules.

\subsection{Input Node Features} \label{subsec:infmod-nodefeats}
At timeslot $j$, the Inference Module receives as input a node feature
matrix $X^{(0)}_j\in \!\mathbb{R}^{N\times D}$ containing the row-ordered node feature
vectors $x_{i, j}^{(0)}\in \mathbb{R}^{D}, \forall i\in V_a$, where $D$
represents the number of features representing a node's input state. 
Since the tags lie in the immediate proximity of a node, and each of them
interacts only with their host, we model them as a feature in their host's
input feature vector. One can also include additional features to assist the
\ac{GNN} during inference.
Hence, we consider three different node features:
\begin{itemize}
    \item Hosted-Tags: the number of tags hosted by a node. 
    \item Node-ID: integer identifying a node in the graph. 
    \item Min. Tag-ID: the minimum tag ID among tags hosted by a node. Since
        a node can host several tags, the min. Tag-ID represents only the
        lowest ID value among its hosted tags. 
\end{itemize}
Intuitively, the number of tags hosted by a node is decisive for assigning
carrier-generating nodes. E.g., if one node hosts all tags, this node should
never be expected to provide the unmodulated carrier in the schedule.
Similarly, the node hosting the greatest number of tags is unlikely to be
a carrier provider in the schedule.  For this reason, \emph{Hosted-Tags}
is always included as an input node feature. Moreover, including the node-ID and the minimum tag-ID can provide the scheduler with context on how to prioritize carrier-provider nodes, and with an order to interrogate the tags.
%

\subsection{Inference Module} \label{subsec:infmod-arch}
\begin{figure}
    \centering
     \includegraphics[width=0.8\linewidth]{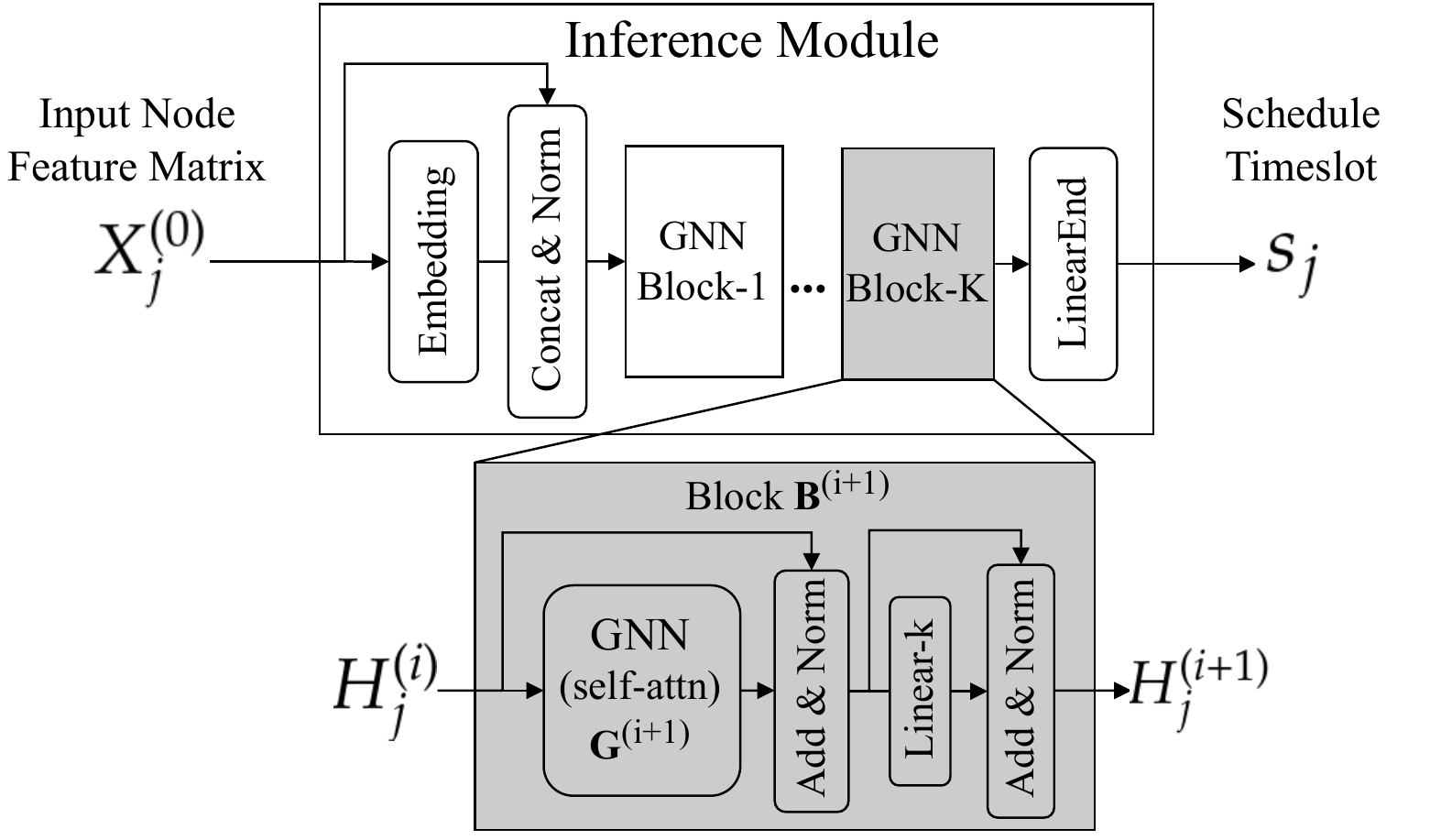}
     \caption{\capt{At the core of the inference module there are $K$
     \acs*{GNN} layers with self-attention.} The Inference Module implements \system's \ac{ML} model.}
     \label{fig:inference-module}
     \vspace*{-0.5cm}
\end{figure}
The Inference Module is the central learning component of \system and contains the
\ac{ML} model for performing inference. 
In the following, we describe in detail the Inference Module's \ac{ML} architectural components.
The Inference Module consists of three parts, as depicted in Figure~\ref{fig:inference-module}. 
%
%

\fakepar{Embedding}
The input node feature matrix $X^{(0)}_j$ is first transformed by an \emph{Embedding} component with the aim to assist the subsequent message-passing operations in performing better injective neighborhood aggregation. That is, enabling each node to better distinguish its neighbors. 
The embedding transformation of $X^{(0)}_j$ is described by:
\begin{equation}
    H^{(0)}_j = \text{LN}\left(\bigparallel \left[X^{(0)}_j, \text{FNN}_e\left( X^{(0)}_j\right) \right] \right) \,\text{,}
    \label{eq:embedding}
\end{equation}
where $\text{LN}$ corresponds to the layer normalization operation from Ba et al.~\cite{Ba2016LayerNorm}, and $\bigparallel$ represents the concatenation operation. $\text{FNN}_e$ corresponds to a fully-connected neural network layer (FCNN) with a non-linear transformation as: 
\setlength{\arraycolsep}{0.0em}
\begin{eqnarray}
    \text{FNN}_e(X^{(0)}_j)&\,=\,&\text{LeLU}\left(\text{FCNN}_e(X^{(0)}_j)\right) \\
    \text{FCNN}_e(X^{(0)}_j) &\,=\,& X^{(0)}_j W_{e} + b_e \,\,\text{,} \label{eq:FCNN}
\end{eqnarray}
\setlength{\arraycolsep}{5pt}
with the leaky ReLU operator $\text{LeLU}$ and NN parameters $W_e\in\mathbb{R}^{D \times D_e}, b_e\in\mathbb{R}^{D_e}$. The embedding outputs the node feature matrix $H^{(0)}_j\in \mathbb{R}^{N\times (D+D_e)}$.  


\fakepar{Stacked GNN Blocks} 
The output from the embedding component
$H^{(0)}_j$ represents the input to a stack
of $K$ \ac{GNN} blocks $\{B^{(i+1)}\}_{i=0}^{K-1}$. The inner structure of each \ac{GNN} block is depicted in the lower part of Figure~\ref{fig:inference-module}.
We stack $K$ different \ac{GNN} blocks to allow nodes to receive input from their $K$-hop transformed neighborhood. 
Moreover, we select summation as the aggregation operation (see Eq.~\ref{eq:main-gnn-update}) motivated by the results of Hamilton et al.~\cite{Hamilton2017inductive}. 
Each \ac{GNN} block $B^{(i+1)}$ receives as input the previous layer output $H^{(i)}_j$ and performs the following operations:
\setlength{\arraycolsep}{0.0em}
\begin{eqnarray}
H^{(i+1)}_j &\,=\,& \textbf{B}^{(i+1)}(H^{(i)}_j) \\
\textbf{B}^{(i+1)}(H^{(i)}_j) &\,=\,& \text{LN}\left(\tilde{H}^{(i+1)}_j + 
\text{FNN}_B^{(i+1)}\left(\tilde{H}^{(i+1)}_j\right) \right) \\ 
\label{eq:block-gnn-NN}
\tilde{H}^{(i+1)}_j &\,=\,& \text{LN}\left(H^{(i)}_j + \textbf{G}^{(i+1)}(H^{(i)}_j)\right) \,\text{.}
\label{eq:block-gnn}
\end{eqnarray}
\setlength{\arraycolsep}{5pt}
Eq.~\ref{eq:block-gnn} corresponds to a multi-head self-attention \ac{GNN} activation $\textbf{G}^{(i+1)}$ followed by an Addition\&Normalization component in Figure~\ref{fig:inference-module}. Subsequently, Eq.~\ref{eq:block-gnn-NN} implements a per-node FCNN with a non-linear transformation, followed by another Add\&Norm component.
Analogous to Eq.~\ref{eq:FCNN}, each block's per-node $\text{FNN}_B^{(i+1)}$ has parameters $W_b^{(i+1)}$ and $b_b^{(i+1)}$ followed by a leaky-ReLU. 

The operation $\textbf{G}^{(i+1)}$ in Eq.~\ref{eq:block-gnn} implements a scaled dot-product multi-head self-attention \ac{GNN}~\cite{Vaswani2017attention, Shia2021transformerconv}. Each of the $N$ row-ordered node feature vectors $H^{(i)}_j=\{h_{j,t}^{(i)}\}_{t=0}^{N-1}$ is updated at \ac{GNN} layer $\textbf{G}^{(i+1)}$ by the following message-passing operation (for simplicity, we omit the timeslot subscript $j$):

\setlength{\arraycolsep}{0.0em}
\begin{eqnarray}
    h_{t}^{(i+1)} & = & \text{FCNN}_{upt}^{(i+1)}\!\!\left(h_{t}^{(i)} \right) + \bigparallel_{m=1}^{M} \left[ \sum_{p\in \mathcal{N}(t)} \alpha^{(i+1)}_{m,tp}\tilde{v}_{m,p}^{(i+1)} \right] \label{eq:main-gnn-update}\\
    \tilde{v}_{m,t}^{(i+1)} &=& \text{FCNN}_{m,val}^{(i+1)} \left(h_{t}^{(i)}\right) \label{eq:neighbor-update}\\ 
    \alpha^{(i+1)}_{m,tp} &=& \frac{\Gamma \left( q_{m,t}^{(i+1)}, k_{m,p}^{(i+1)}\right)}{\sum\limits_{u\in \mathcal{N}(t)} \Gamma \left( q_{m,t}^{(i+1)}, k_{m,u}^{(i+1)}\right)} \label{eq:att-weights}\\
    q_{m,t}^{(i+1)} &=& \text{FCNN}_{m,qry}^{(i+1)}\!\left(h_{t}^{(i)}\right) \label{eq:query}\\ 
    k_{m,t}^{(i+1)} &=& \text{FCNN}_{m,key}^{(i+1)}\!\left(h_{t}^{(i)}\right) \label{eq:key}\\ 
    \Gamma \left( A, B\right) &=&\,\, \frac{\exp\left( A^{\top} B\right)}{\sqrt{d}} \,\text{,}\,\, A,B \in \mathbb{R}^{d} \,\,\,\text{.} \label{eq:scaled-dot}
\end{eqnarray}
\setlength{\arraycolsep}{5pt}
Eq.~\ref{eq:main-gnn-update} is the main node update operation for each node, consisting of adding two parts. First, a transformation of the target node's feature vector through $\text{FCNN}_{upt}^{(i+1)}$. Second, the concatenated $M$-heads of the attention operations with the target node's neighbors.
Each self-attention operation $m$ consists of a weighted sum over transformed neighboring node feature vectors $\tilde{v}_{m,t}^{(i+1)}$, inspired by the attention operations by Vaswani et al.~\cite{Vaswani2017attention}.  Each of the neighbors' node feature vectors are transformed by the layer $\text{FCNN}_{m,val}^{(i+1)}$ in Eq.~\ref{eq:neighbor-update}. The weights $\alpha^{(i+1)}_{m,tp}$ at attention head $m$ from neighbor source $p$ to target update node $t$ are obtained through Eq.~\ref{eq:att-weights} over the scaled dot product operation $\Gamma$ (Eq.~\ref{eq:scaled-dot}) between query $q_{m,t}^{(i+1)}$ and key $k_{m,t}^{(i+1)}$ transformations of the target node (see Eq.~\ref{eq:query}) and the source neighbor node (see Eq.~\ref{eq:key}), respectively. 
%

There are multiple reasons motivating the use of
attention. 
First, 
it allows to leverage neighboring nodes' contributions differently.
Second, 
attention has provided better results for node classification tasks~\cite{Velickovic2018gat} compared to isotropic \acp{GNN}~\cite{Kipf2017gcn, Hamilton2017inductive}. 
Moreover,
conventional \ac{GNN} node classification benchmarks assume that nodes in a spatial locality of the graph assume similar labels and leverage the fixed-point theorem property in stacking \acp{GNN}~\cite{Kipf2017gcn}. This greatly contrasts with the carrier scheduling problem, in which neighboring nodes are mostly expected to have different classes (a node provides a carrier, neighbors interrogate tags). Finally, attention serves as a counteracting factor for this property when combined with skip-connections: it builds deeper networks that can
learn contrasting representations at each layer.

\fakepar{Classification Layer} The output from the $K$ \ac{GNN} blocks is then fed to a FCNN followed by a Softmax for obtaining a per-node probability distribution over the classes that correspond to the possible node actions $\{\mathtt{C}, \mathtt{T}, \mathtt{O}\}$. Finally, each node is assigned to the class with the highest probability.

Note that all parameters from the Inference Module listed (Eqs. \ref{eq:embedding}--\ref{eq:key}) are shared across timeslots $s_j$.

\section{Learning to Schedule} \label{sec:learning-to-schedule}
\begin{table*}
    \centering
    \caption{
    \capt{The interrelation between data generation and input node features strongly impact the model's performance. Including symmetry-breaking constraints dramatically improves the model performance.} Interdependence of data generation and input feature configuration for an Inference
    Module of six GNN-Blocks. $S_{corr}$ indicates percentage of problem instances for which the NN delivers a complete schedule. F1-score is provided for $\mathtt{C}$ (carrier class).
    }
    \label{tab:Ml-alg-perf}
    \begin{tabular}{@{} l *{7}{c} }
    \toprule
        \multicolumn{2}{r}{Symmetry breaking:} & \multicolumn{3}{c}{Disabled} & \multicolumn{3}{c}{Enabled} \\
        \cmidrule(lr){3-5} \cmidrule(l){6-8}
        
        \multicolumn{2}{r}{Performance Metric [\%]:} & Accuracy & F1-score & $S_{corr}$ & Accuracy & F1-score & $S_{corr}$  \\
        \midrule
        \multicolumn{2}{r}{\emph{Features-1} \small{(Hosted-Tags)}:} & 85.69 & 57.78 & 27.04     & 86.61 & 60.54 & 10.11 \\
        \multicolumn{2}{r}{\emph{Features-2} \small{(Hosted-Tags + Node-ID + Min. Tag-ID.)}:} & 86.40 & 59.82 & 47.96     & \textbf{99.22} & \textbf{97.36} & \textbf{99.64} \\
        \bottomrule 
    \end{tabular}
\end{table*} 
This section explores how the Inference Module's ability to compute
interrogation schedules is influenced by two factors. On one hand, the
interdependence between the choice of input node features and on the other the configuration
of the training data generation. Additionally, we explore the influence of
\ac{ML} model complexity in terms of the number of \ac{GNN} blocks on the Inference
Module's performance.  This exploratory analysis resulted in selecting an
Inference Module composed of 12 \ac{GNN} blocks. 

\subsection{Training Data Generation}

The carrier scheduling problem, as described in
Eqs.~\ref{eq:optobj}-\ref{eq:optconst2}, 
presents
symmetries that result in multiple optimal solutions which confuse the
\ac{DL} model while training. In the following, we describe how we leverage
symmetry-breaking constraints in the constraint optimizer to generate a
training dataset with unique optimal solutions.

As an example of these symmetries consider that: because
the order of the timeslots in the schedule is irrelevant, a schedule of
duration $L$ is equivalent to $L!$ other schedules. 
A similar set of symmetries
appears among all tags hosted by the same node as the order of interrogating
them is also irrelevant. To make matters worse, a given set of tags can often
be served by more than one carrier generator, therefore introducing more
symmetries. 

\fakepar{Symmetry-Breaking Constraints} 
To overcome the confusion that these symmetries might cause to a learning-based model
during training, we further constrain the carrier scheduling problem in a way
that eliminates symmetries but does not otherwise alter the problem.
To that end, we enforce lexicographical minimization of a vector of length $T$
that indicates the timeslot where each tag is scheduled. This automatically
eliminates symmetries related to the order of tag interrogations.
Similarly, we also lexicographically minimize another length-$T$ vector
containing the node that provides the carrier for each tag; which eliminates
symmetries related to multiple potential carrier nodes.
\begin{figure}
    \centering
    \includegraphics[width=0.92\linewidth]{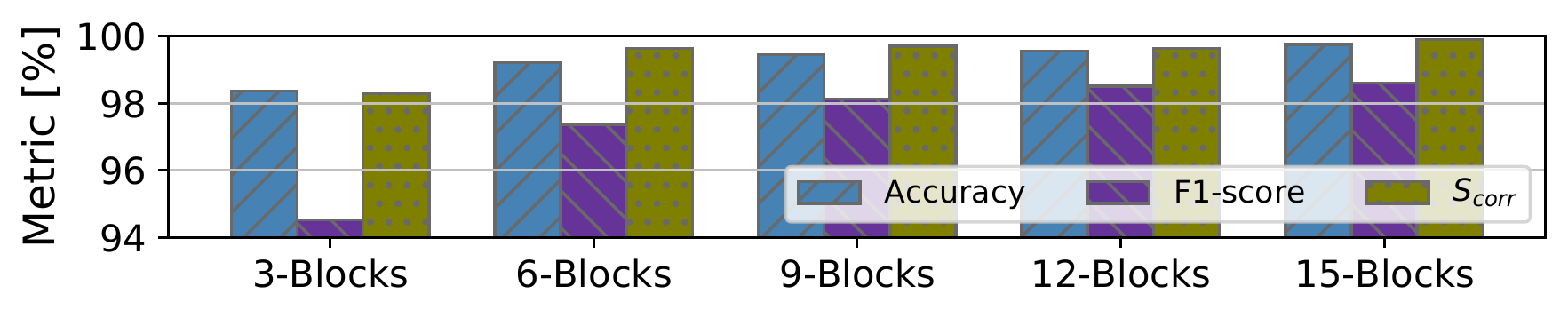}
    \caption{\capt{Increasing the number of GNN-Blocks improves the performance
        of the NN model.} The NN reaches a saturation point for the F1-score at approx. 12 layers.}
    \label{fig:layers-analysis}
    \vspace*{-0.5cm}
\end{figure}
\subsection{\ac{ML} Model Validation Setup}
We implement all components in the Inference Module using PyTorch~\cite{pytorch2019}. For the \acp{GNN} layers $\textbf{G}^{(i+1)}$, we use the PyG self-attention based GNN \emph{TransformerConv} implementation~\cite{Fey2019torchgeo}. 

\fakepar{Input Features} We consider two different feature configurations for the input node feature matrix $X_1^{(0)}\in\mathbb{R}^{N\times D}$: \emph{Features-1} includes only the \emph{Hosted-Tags} ($D=1$), and \emph{Features-2} considers Features-1 plus the \emph{Node-ID} and \emph{minimum Tag-ID} among the tags hosted by a node ($D=3$). 

\fakepar{Validation Dataset} We generate small-sized problem instances of
varying sizes and number of tags
from two to ten nodes ($N\in[2, 10]$), and hosting one to 14 tags ($T\in[1, 14]$). We
generate these graphs using the \emph{random geometric} graph generator from
NetworkX~\cite{NetworkX2008} to guarantee that these graphs can exist in 3D space.
As the network size varies, we make sure to maintain a constant network spatial density. 
We uniformly assign tags to hosts at random. We employ a constraint optimizer to compute
solutions for a total of 520000 problem instances both including and without
including the symmetry-breaking constraints. As constraint optimizers, we
employ MiniZinc~\cite{Nethercote2007minizinc} and OR-Tools~\cite{ortools2019}. 
The problem instances are divided into a train-set and a validation-set using a 80\%-20\% split. 
Since we perform a per-node classification for every scheduling timeslot (see Figure~\ref{fig:pred-flow}), we further consider each timeslot input-target pair $(X_j^{(0)}, s_j)$ as a training sample, which yields an approximate total of 1.5 million samples.

\fakepar{Validation Metrics} We consider both \ac{ML} metrics and an application related metric to evaluate a model's performance. 
From the ML perspective, we employ overall accuracy, and the carrier class ($\mathtt{C}$) F1-score due to its crucial role in avoiding signal interference for tag interrogation. 
For the application-related metric we consider the percentage of correctly computed schedules $S_{corr}$. This metric indicates the percentage of problem instances for which the
already-trained ML model produces a complete (all timeslots) and correct
(fulfilling carrier scheduling constraints) schedule. 
At inference time, the \emph{Topology Handler} handles these unlikely cases
as described in Section~\ref{subsec:sysarch}.

\subsection{ML Model Training} \label{subsec:training}
We train the Inference Module with the 
Adam optimizer~\cite{Kingma2015adam} on the basis of mini-batch gradient descent with standard optimizer parameters and an initial learning rate of $10^{-3}$.
The \ac{ML} model should give greater importance to the carrier-generating node class ($\mathtt{C}$) since it can subsequently determine the tags that can be interrogated or not due to signal interference. 
Hence, we build upon the cross-entropy loss for classification and propose an additional factor to account for greater importance to the carrier class ($\mathtt{C}$):
\setlength{\arraycolsep}{0.0em}
\begin{eqnarray}
    L_{batch} &=& \frac{1}{N_b} \sum_{\hat{y}=0}^{N_b-1} L_x(\hat{y}, y) \,\, L_1(\hat{y}, y) + \rho||\hat{W}||_2^2 \\
    L_x(\hat{y}, y) &=& -\!\!\!\!\!\sum_{k\in\{\mathtt{C}, \mathtt{T/O}\}}\!\!\! y_k \log(\hat{y}_k)  \\
    L_1(\hat{y}, y) &=& \exp\left( ||(\hat{y} == \mathtt{C}) - (y == \mathtt{C})|| \right) \, \text{,}
\end{eqnarray}
\setlength{\arraycolsep}{5pt}
where $N_b$ is the number of nodes in the mini-batch, $\rho$ is the regularization hyperparameter, and $\hat{W}$ represents the tensor containing all the learning parameters undergoing gradient descent. We implement learning rate decay by $2\%$ every epoch, and early stopping after 25 subsequent epochs without minimization of the test loss, and save the best-performing model on the basis of the F1-score. 
\subsection{Validation Results}
The Inference Module considered for evaluating data generation strategies and node feature representation consists of six GNN-blocks. We empirically chose the number of GNN-blocks for the first experiments based on a trade-off between model training time and performance. 
After establishing the best combination of input node feature representation and data generation strategy, we analyze the influence in performance of \ac{ML} model complexity.

\subsubsection{Data Generation and Input Features Interdependence}
The influence of both input node feature representation and data generation configuration in model performance is depicted in Table~\ref{tab:Ml-alg-perf}.
Regardless of the data-generating configuration, an enriched node feature representation (including more node features) leads to an increase in accuracy and F1-score. There are two possible reasons for this. First, by increasing the node feature dimension $D$, we ensure that the GNNs can perform more efficient injective neighborhood aggregation (to better distinguish neighboring node contributions). This is realized by making it less likely for two nodes to have the same input feature vector to the \ac{GNN}. Second, both the Nodes-ID and Tags-ID feature serve as a node's positional encoding information in a graph, a property that assists the model in breaking graph symmetries~\cite{You2019positionaware, Dwivedi2020benchmarking}. 

Additionally, an enriched node feature representation leads  to an
increase of $S_{corr}$, regardless of the chosen data generation configuration: $20.92\%$ and $89.53\%$ improvement from Features-1 to Features-2 for the standard and the symmetry-breaking configurations, respectively.
The constraint optimizer explicitly uses the Node-ID and Tag-ID to compute the optimal solution in the symmetry-breaking configuration, which is why providing the \ac{ML} model with these features is crucial for it to be able to learn the structural dependencies in the topologies. 
Implementing symmetry-breaking measurements in the data generation procedure is a critical measure for allowing the NN model to generate complete schedules (highest increase in $S_{corr}$). Since we explore a supervised learning approach, it is crucial to constrain the mapping between inputs and targets for the NN model to learn consistent graph-related structural dependencies.

\subsubsection{Influence of the Number of GNN-Blocks}
\label{subsec:layereval}
To analyze the influence of the required $K$-hop neighborhood aggregation of the \ac{GNN} model, we analyze the model's performance as we vary the number of GNN-Blocks while keeping other architectural components constant to the values found by extensive empirical analysis. 
Specifically, we set the embedding dimension to $D_e=48$, the number of attention heads to $M=2$, and the GNN-Blocks' hidden feature dimensions to $D_H^{(i)}=200$.
Figure~\ref{fig:layers-analysis} illustrates how increasing the number of
GNN-Blocks increases the performance of the NN model: 
the F1-score increases to a saturation point around 12 layers. Although the
15-layer model exhibits the highest $S_{corr}$, subsequent experiments showed that
this model overfits and hence is unable to scale to larger problem instances.

\fakepar{Inference Module} Based the findings of this section, \system's Inference Module consists of 12 \ac{GNN} blocks and is trained on $\sim\!\!\!424000$ small-sized problem instances (\ac{IoT} networks of up to 10 nodes and 14 tags) obtained from the optimal scheduler as described in Section~\ref{subsec:training}. We used an NVIDIA Titan RTX for $131$ epochs before reaching the early-stop condition.
The model achieved $99.56\%$ accuracy, a carrier-class F1-score of $98.51\%$, and a percentage of correctly computed schedules of $S_{corr}=99.88\%$.

\section{Evaluation}
\label{sec:evaluation}

After designing \system's Inference Module in Section~\ref{sec:learning-to-schedule}, in
this section, we compare \system's results to those of
the TagAlong scheduler~\cite{PerezPenichet2020afast} and the optimal scheduler.
We highlight the following key findings:
\begin{itemize}
    \item \system performs within 3\% of the optimal scheduler on the average number
        of carriers used, while consistently outperforming TagAlong by up to 50\%.
        This directly translates into energy and spectrum savings for the \ac{IoT} network.
    \item Our scheduler scales far beyond the problem sizes where it was
        trained, while still outperforming TagAlong; therefore enabling large
        resource savings well beyond the limits of the optimal scheduler.
    \item We deploy \system to compute schedules for a real \ac{IoT} network of 24 nodes. Compared to the TagAlong scheduler, \system reduces the energy per tag interrogation by $13.1\%$ in average and up to $51.6\%$.
    \item With polynomial time complexity and a maximum observed computation time of \SI{1.49}{\s}, \system's speed is comparable to TagAlong's, and  well within the needs of a practical deployment.
\end{itemize}

\fakepar{Baselines} 
To conduct our evaluation, 
we consider two baselines: i) the \textit{optimal scheduler}, which corresponds to using a constraint optimizer for computing the optimal solution including the symmetry-breaking constraints for topologies of up to $10$ nodes and $14$ tags, and ii) the
\textit{TagAlong scheduler}, the state-of-the-art heuristic algorithm for computing interrogation
schedules for the scope of backscatter \ac{IoT} networks considered in this work~\cite{PerezPenichet2020afast}.

\fakepar{Evaluation Section Structure} This section is structured as follows. Sec.~\ref{subsec:test-set-perf} benchmarks \system's performance against the optimal scheduler and the TagAlong scheduler for topologies of up to $10$ nodes and $14$ tags previously unseen to \system (\emph{test} set). Sec.~\ref{subsec:generalization-perf} analyzes \system's gains over TagAlong for topologies well beyond the practical applicability of the optimal scheduler of up to $60$ nodes and $160$ tags (\emph{generalization} set). Sec.~\ref{subsec:computation-time} then briefly describes \system's time complexity and its computation times. Finally, Sec.~\ref{subsec:real-network-perf} demonstrates \system's ability to produce schedules for a real \ac{IoT} network of $N=24$ nodes under different number of tags configurations.

\subsection{Test Set Performance} \label{subsec:test-set-perf}

\begin{figure}
    \centering
    \begin{subfigure}[b]{0.48\linewidth}
        \includegraphics[width=\linewidth]{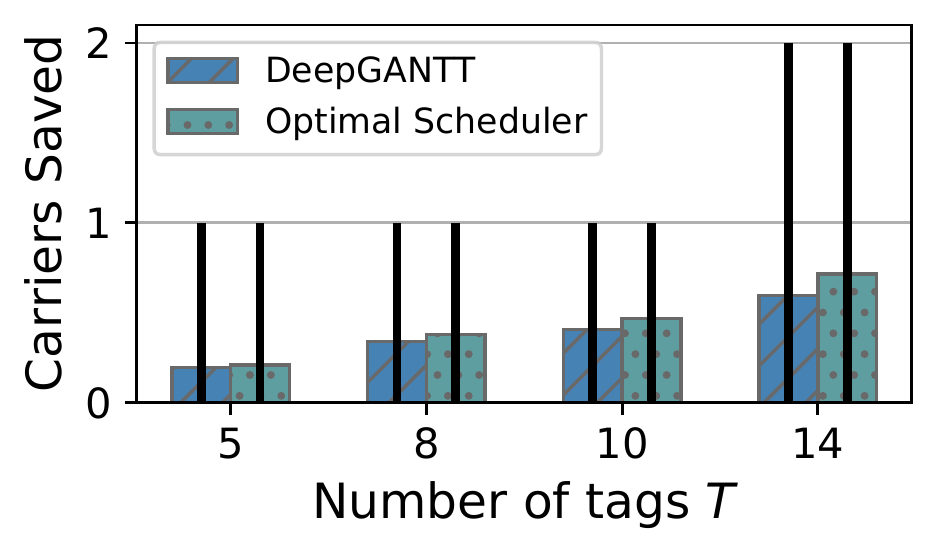}
        \caption{\capt{Average number of carriers saved relative to TagAlong}.}
        \label{fig:testset-aggregated-gains}
    \end{subfigure}
    \hfill
    \begin{subfigure}[b]{0.48\linewidth}
        \includegraphics[width=\linewidth]{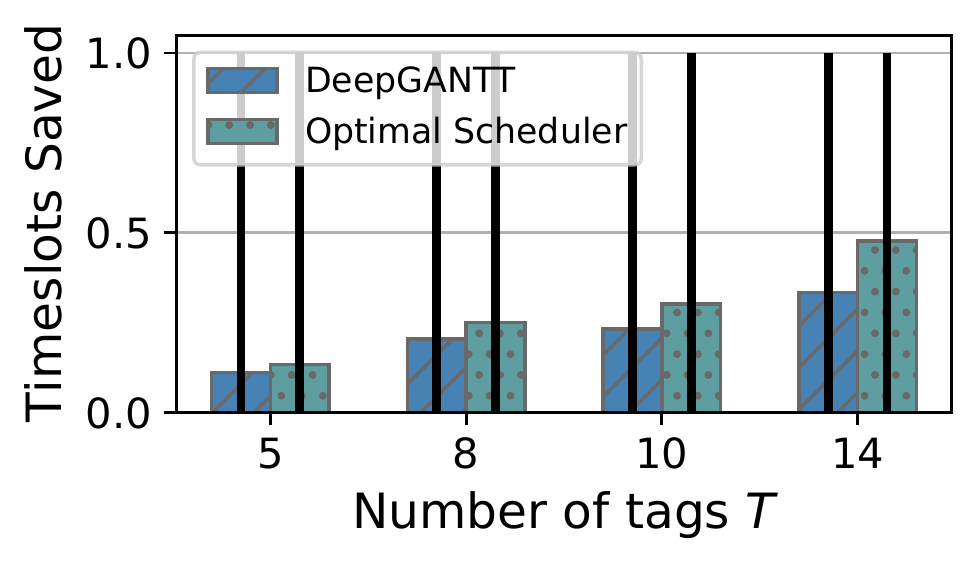}
        \caption{\capt{Average timeslot savings relative to TagAlong}.}
    \label{fig:timeslots-saved}
    \end{subfigure}
    \caption{\capt{\system largely mimics the optimal scheduler both in terms of carriers saved and timeslots saved}. Average gain for networks of 10 IoT nodes and various numbers of tags in the test dataset. The black bars depict the 10 and 90 percentiles.}
    \label{fig:testset-eval}
    \vspace*{-0.4cm}
\end{figure}

We hereby demonstrate \system's ability to mimic the behaviour of the optimal scheduler to produce interrogation schedules and exhibit similar gains as the optimal scheduler's performance over the TagAlong heuristic.

\fakepar{Test Dataset} We consider topologies consisting of $\sim\!\!106000$ problem instances of up to $10$ nodes and $14$ tags for which it is still possible to deploy the optimal scheduler.

\fakepar{Test Metrics} 
We compare the number of carrier slots in the generated schedules since it
directly affects the \ac{IoT} network's overall energy consumption. Specifically, for every
evaluation problem instance we compute \system's saved carriers as $C_d - C_t$
and, those of the optimal scheduler as $C_o - C_t$; where $C_d$, $C_t$, and
$C_o$ are the number of carriers scheduled by the \system, TagAlong and the optimal
schedulers respectively.
Furthermore, we analyze the total schedule length, since it directly relates to
the latency of communications in the wireless network. We compute the timeslots savings in a manner analogous to the carrier savings.

\fakepar{Test Results}
Figure~\ref{fig:testset-aggregated-gains} shows the average number of carriers
saved compared to TagAlong (higher is better) for various network sizes in the test dataset. \system's performance 
is very close to that of the optimal scheduler in
all cases.
The number of timeslots in the schedule is the secondary objective in the tag
scheduling problem; this is because, while the duration of the schedule can
impact latency and other performance metrics in the network, the number of
carriers directly impacts energy and spectral efficiency. Note that reducing the
number of carrier slots, can potentially  also shorten the schedule owing to
carrier reuse~\cite{PerezPenichet2020afast}. 
As depicted in Figure~\ref{fig:timeslots-saved}, \system largely mimics the performance of the optimal scheduler regarding
timeslot savings. Our scheduler outperforms the TagAlong scheduler in $98.2\%$ of the test-set instances.
On those instances where TagAlong schedules are shorter, it is by one timeslot at most.

\subsection{Generalization Performance} \label{subsec:generalization-perf}

\begin{figure}
    \centering
    \begin{subfigure}[b]{0.48\linewidth}
        \includegraphics[width=\linewidth]{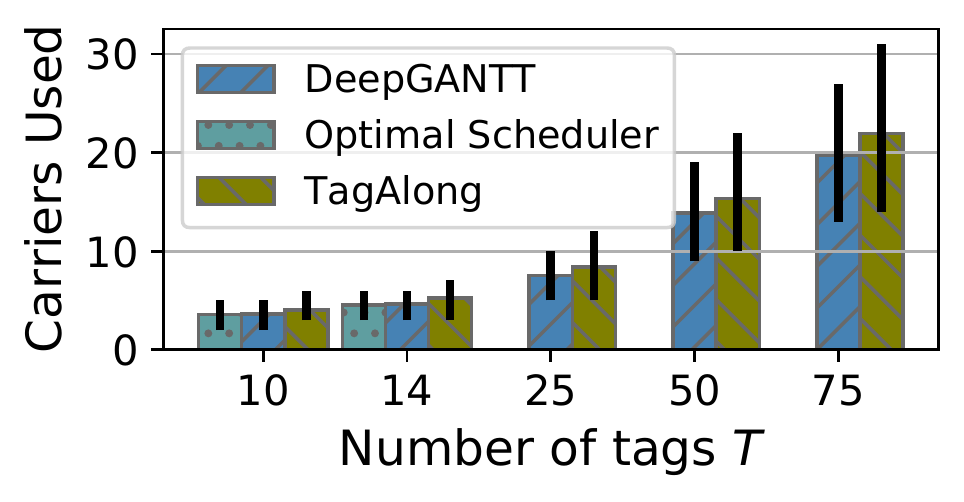}
        \caption{\capt{Average number of carriers used for networks of 10 \ac{IoT} nodes}.}
        \label{fig:testset-carriers-saved-vs-tags}
    \end{subfigure}
    \hfill
    \begin{subfigure}[b]{0.48\linewidth}
        \includegraphics[width=\linewidth]{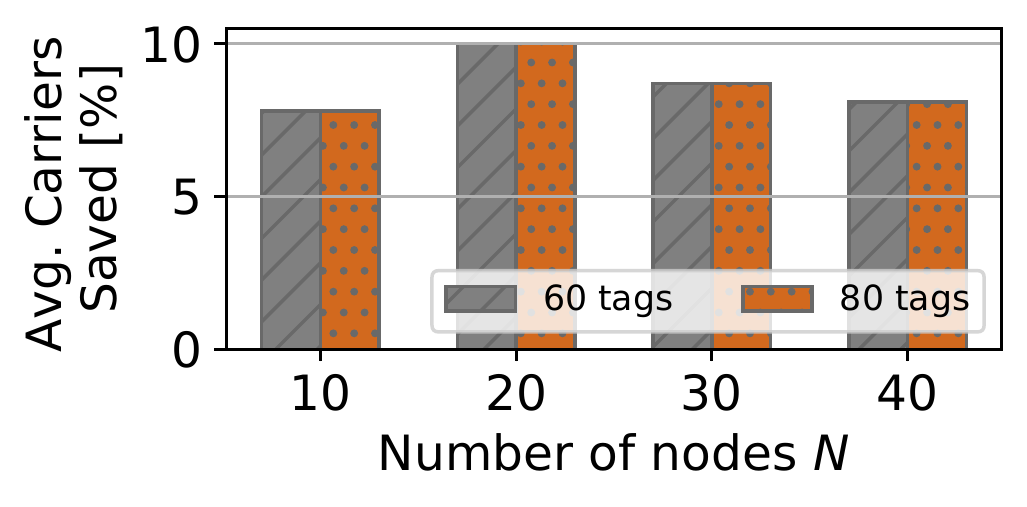}
        \caption{\capt{Average percentage of carriers saved compared to
    TagAlong.}} 
    \label{fig:scalability-carriers-saved}
    \end{subfigure}
    \caption{\capt{\system scales far beyond the range of data where the optimal schedules used in training are available, while 
        outperforming TagAlong by a growing margin (\ref{fig:testset-carriers-saved-vs-tags})} The black bars depict the 10 and 90 percentiles. \system also maintains an average saving of almost 10\% in
        scheduled carriers while scaling up to four times the maximum
    training network size (\ref{fig:scalability-carriers-saved}).}
    \label{fig:generalization-eval-small}
    \vspace*{-1.0cm}
\end{figure}

\begin{figure*}
     \centering
     \begin{subfigure}[b]{0.245\linewidth}
         \centering
         \includegraphics[width=0.98\linewidth]{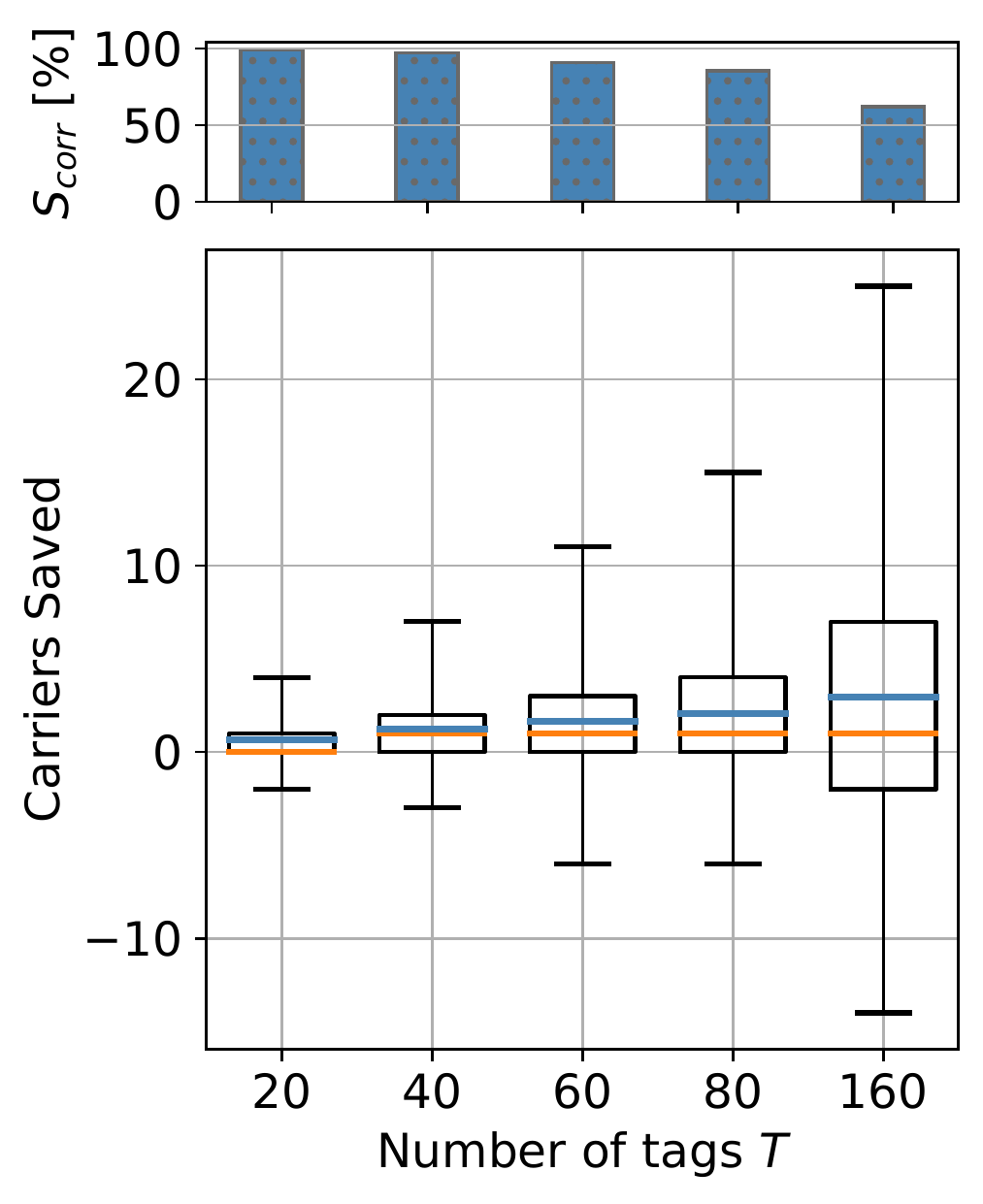}
         \caption{10 node topologies.}
         \label{subfig:sched-seq-10}
     \end{subfigure}
     \hfill
     \begin{subfigure}[b]{0.245\linewidth}
         \centering
         \includegraphics[width=0.98\linewidth]{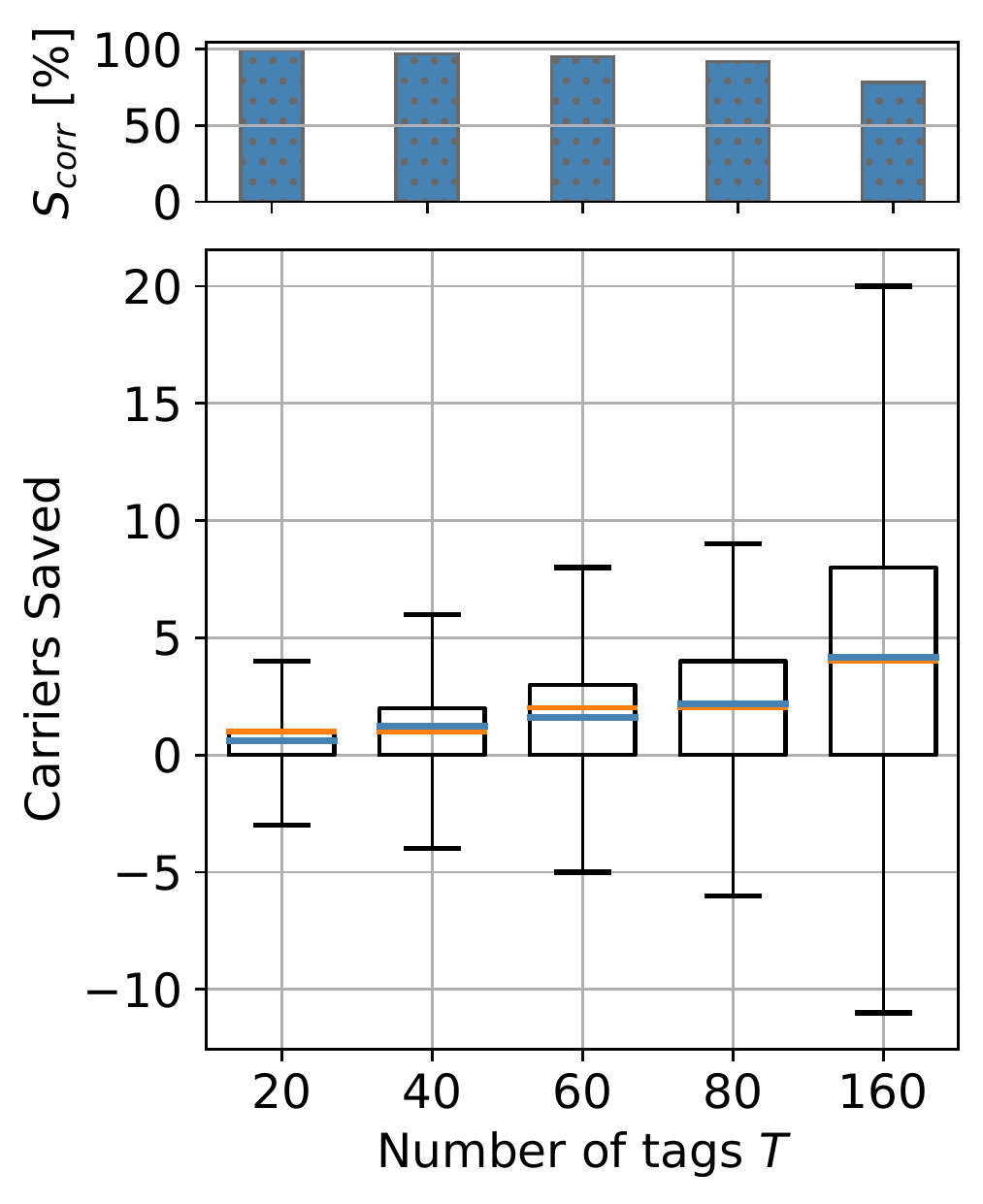}
         \caption{20-node topologies.}
         \label{subfig:sched-seq-20}
     \end{subfigure}
     \hfill
     \begin{subfigure}[b]{0.245\linewidth}
         \centering
         \includegraphics[width=0.98\linewidth]{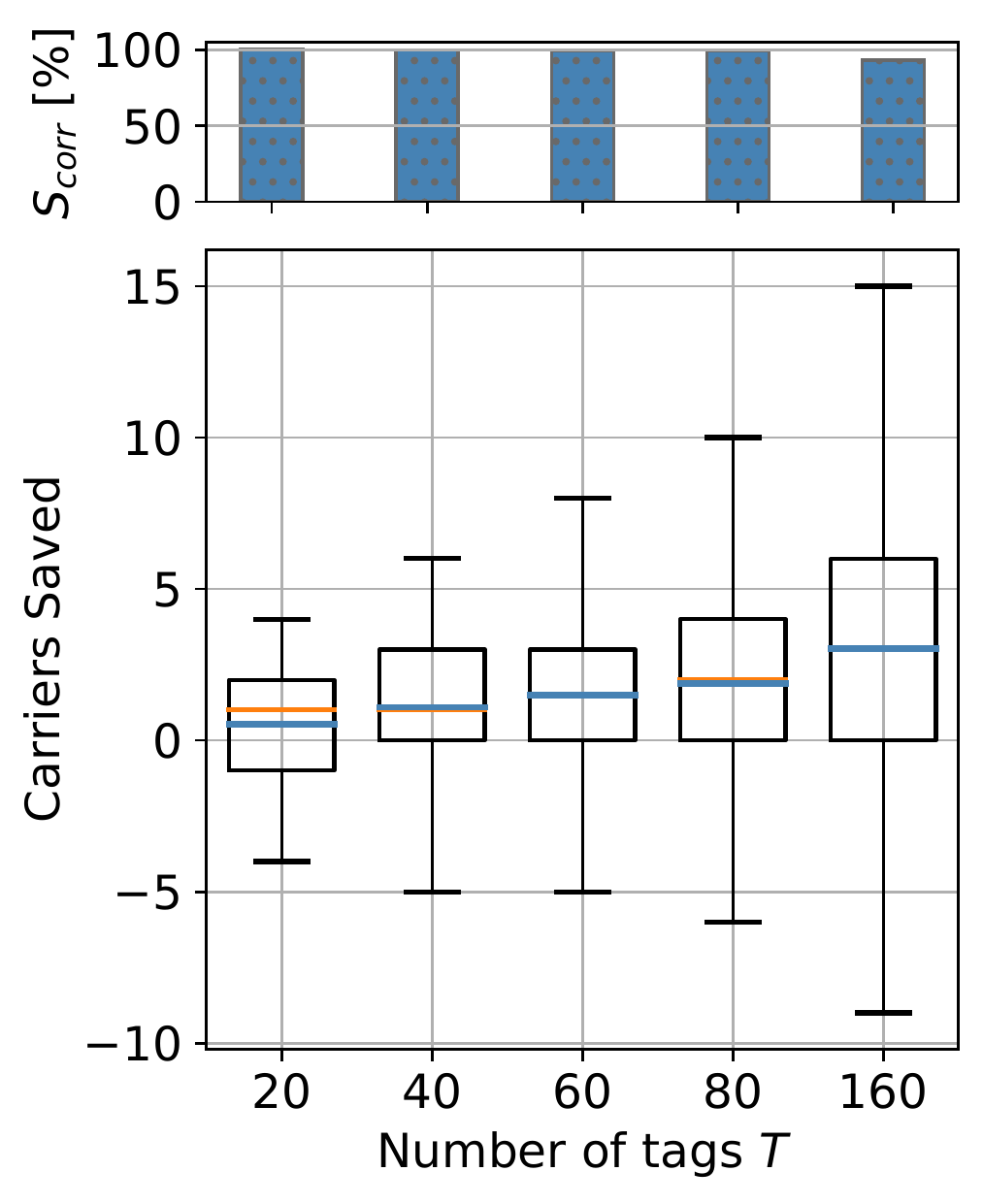}
         \caption{40-node topologies.}
         \label{subfig:sched-seq-40}
     \end{subfigure}
     \hfill
     \begin{subfigure}[b]{0.245\linewidth}
         \centering
         \includegraphics[width=0.98\linewidth]{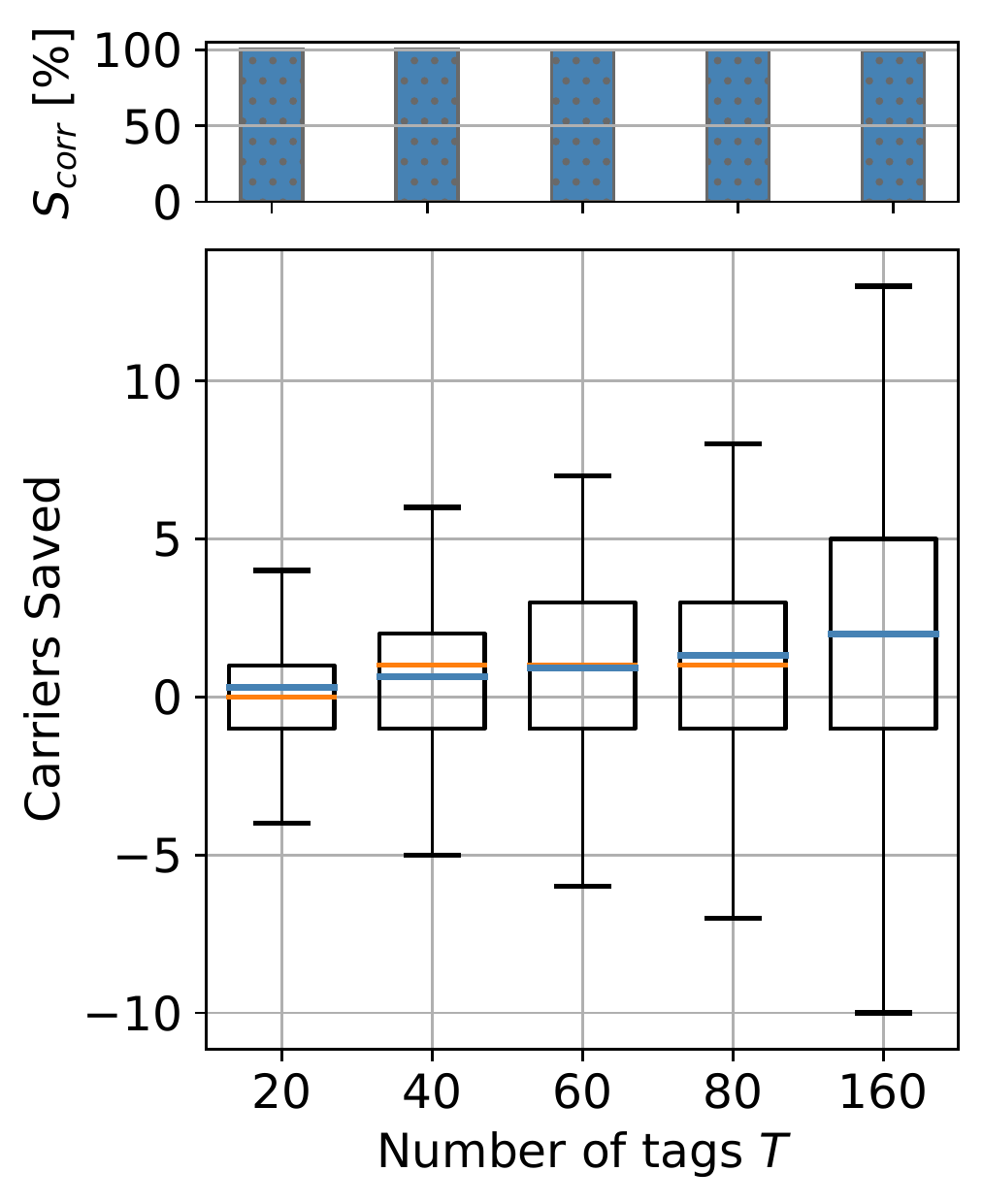}
         \caption{60-node topologies.}
         \label{subfig:sched-seq-60}
     \end{subfigure}
    \caption{
        \capt{\system outperforms TagAlong even when increasing the topology
        sizes far beyond those seen on training.} Scaling capabilities of
        \system when compared to the TagAlong heuristic in terms of carrier
        savings.
        The model achieves
    a maximum carrier reduction of up to \textit{$43.42\%$, $43.86\%$, $33.93\%$,
    $32.14\%$} for 160 tags and $10$, $20$, $40$, and $60$ nodes, respectively. The
    blue and the orange line represent the mean and the median, respectively. Box
    extents delimit 25 and 75 percentiles. Whiskers delimit 1 and 99
    percentiles. The model also has a high success rate ($S_{corr}$), especially for
    larger topologies.}
    \label{fig:scalability-all}
\end{figure*}
We now analyze the capabilities of \system in computing schedules for problem
sizes well beyond those observed during training and compare its performance against the TagAlong scheduler. The ability to generalize
this way directly translates into better scalability than that of the optimal
scheduler.

\fakepar{Generalization Dataset} We consider $1000$ problem instances for every $(N, T)$ pairs from the sets $N\in\{10, 20, 30, 40, 60\}$ and $T\in\{20, 40, 60, 80, 160\}$, i.e., $25000$ different \ac{IoT} wireless networks.

\fakepar{Metrics} We consider the same metrics as those used in Sec.~\ref{subsec:test-set-perf}.

\fakepar{Generalization Results} 
Figure~\ref{fig:testset-carriers-saved-vs-tags} shows a comparison of the average
number of carriers utilized (lower is better) on problem sizes beyond those used
in training. 
\system outperforms TagAlong by a
growing margin well beyond the maximum size of optimal solutions seen in training (beyond $10$ nodes and $14$ tags). Compared to
TagAlong, \system is able to reduce the percentage of necessary carriers 
$\sim\!\SIrange[range-phrase = -]{7}{10}{\percent}$ on average for large numbers of tags, as depicted in
Figure~\ref{fig:scalability-carriers-saved}.

Figure~\ref{fig:scalability-all} depicts the number of carriers saved and the
percentage of correctly computed schedules $S_{corr}$ for different network size
configurations. 
\system consistently increases
the mean number of carriers saved as the number of tags increases for all
considered configurations. 
While there are cases where TagAlong outperforms \system, these are actually
rare occurrences; this is evident by the positive mean and the location of the 25-percentiles. 
Additionally, \system's percentage of correctly-computed schedules ($S_{corr}$)
decreases for the 10-node 80-tags case, but
remains above $99\%$ for the 30, 40 and 60 nodes problem instances, for all
the number of tags considered.
\system is able to reduce
the number of carriers by up to $50\%$ for all number of nodes configurations in
Figure~\ref{fig:scalability-all}. 

\subsection{Computation Time} \label{subsec:computation-time}
\begin{figure}
    \centering
    \begin{subfigure}[b]{0.48\linewidth}
        \includegraphics[width=\linewidth]{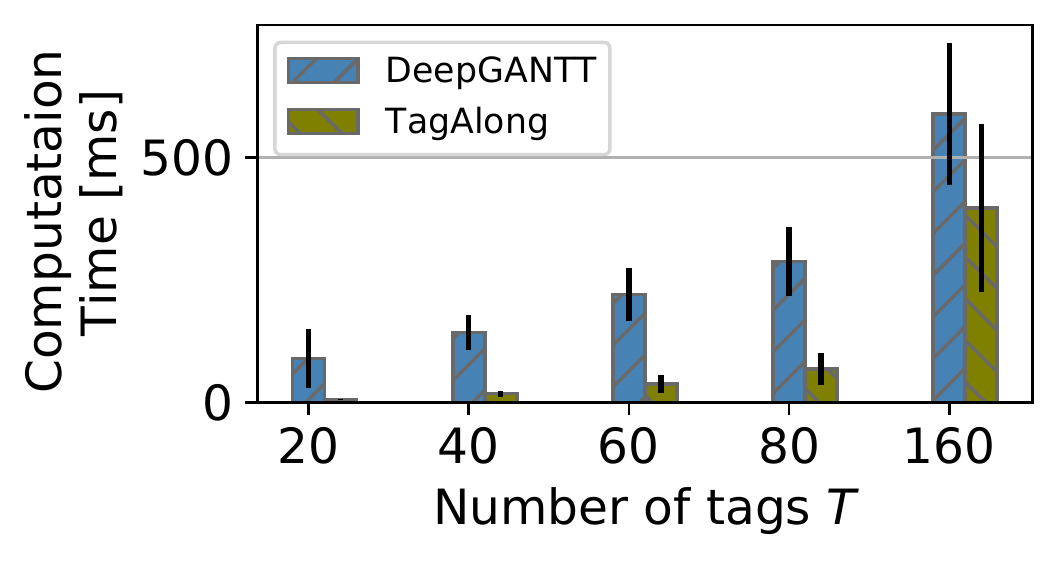}
        \caption{\capt{}Average runtimes for 10 Nodes.}
        \label{fig:runtime-10nodes}
    \end{subfigure}
    \hfill
    \begin{subfigure}[b]{0.48\linewidth}
        \includegraphics[width=\linewidth]{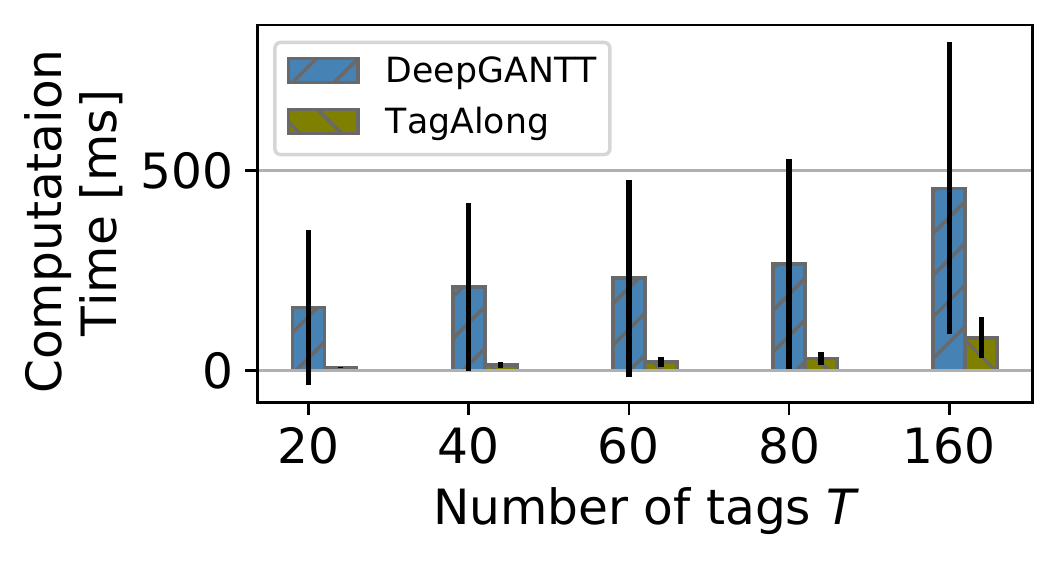}
        \caption{\capt{}Average runtimes for 60 Nodes.} 
    \label{fig:runtime-60nodes}
    \end{subfigure}
    \caption{\capt{While TagAlong runs faster than \system, both run times are
        so small that the difference is negligible.} Run time comparison of
        \system and TagAlong for 10 and 60~\acs*{IoT} nodes with various numbers of
        tags. The black bars represent the standard deviation.}
    \label{fig:runtime}
    \vspace*{-0.5cm}
\end{figure}

A determining factor for the real-world applicability of a scheduler is
the computation time. Figure~\ref{fig:runtime} depicts a run time
comparison of \system and TagAlong on the same hardware.  While TagAlong runs faster, the absolute
values are so small that the difference is negligible in practice.  
Note that for the largest problem instances considered ($40$ nodes
and $80$ tags), the maximum runtime recorded was \SI{1.49}{\s}, with an
average runtime of \SI{0.25}{\s}. This is in stark contrast with the optimal
scheduler that takes
several hours to compute schedules for just 10 nodes and 14
tags.

\fakepar{Time Complexity} 
In general, an attention-based message passing operation
has time complexity $\mathcal{O}(N\!+\!|E|)$, where $N$ is the number of nodes in
the graph and $|E|$ is the number of edges~\cite{Velickovic2018gat}. In the
worst case, \system performs $T$ complete \ac{ML} model passes, one for every tag in the
wireless network. Hence, the complexity of our algorithm is polynomial in
the input size: $\mathcal{O}(T(N\!+\!|E|))$.

\subsection{Performance on a Real \ac{IoT} Network} \label{subsec:real-network-perf}

In this section we deploy \system to compute tag interrogation schedules for a real \ac{IoT} network and compare its performance with that of the TagAlong scheduler. We show that \system can achieve up to $51.6\%$ energy savings in tag interrogation.

\begin{figure}
     \centering
     \begin{subfigure}[b]{0.38\linewidth}
         \centering
         \includegraphics[width=0.98\linewidth]{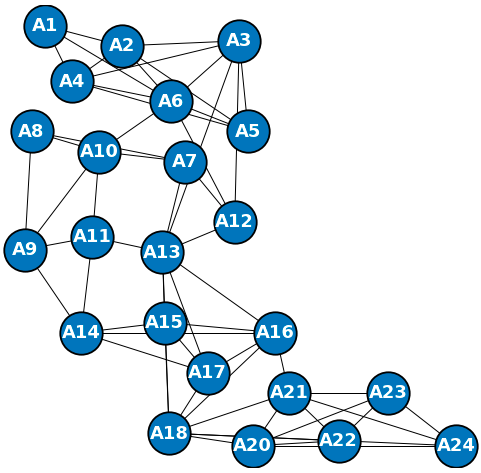}
         \caption{\capt{IoT network topology}.}
         \label{subfig:pi-testbed-topo}
     \end{subfigure}
     \hfill
     \begin{subfigure}[b]{0.6\linewidth}
         \centering
         \includegraphics[width=0.98\linewidth]{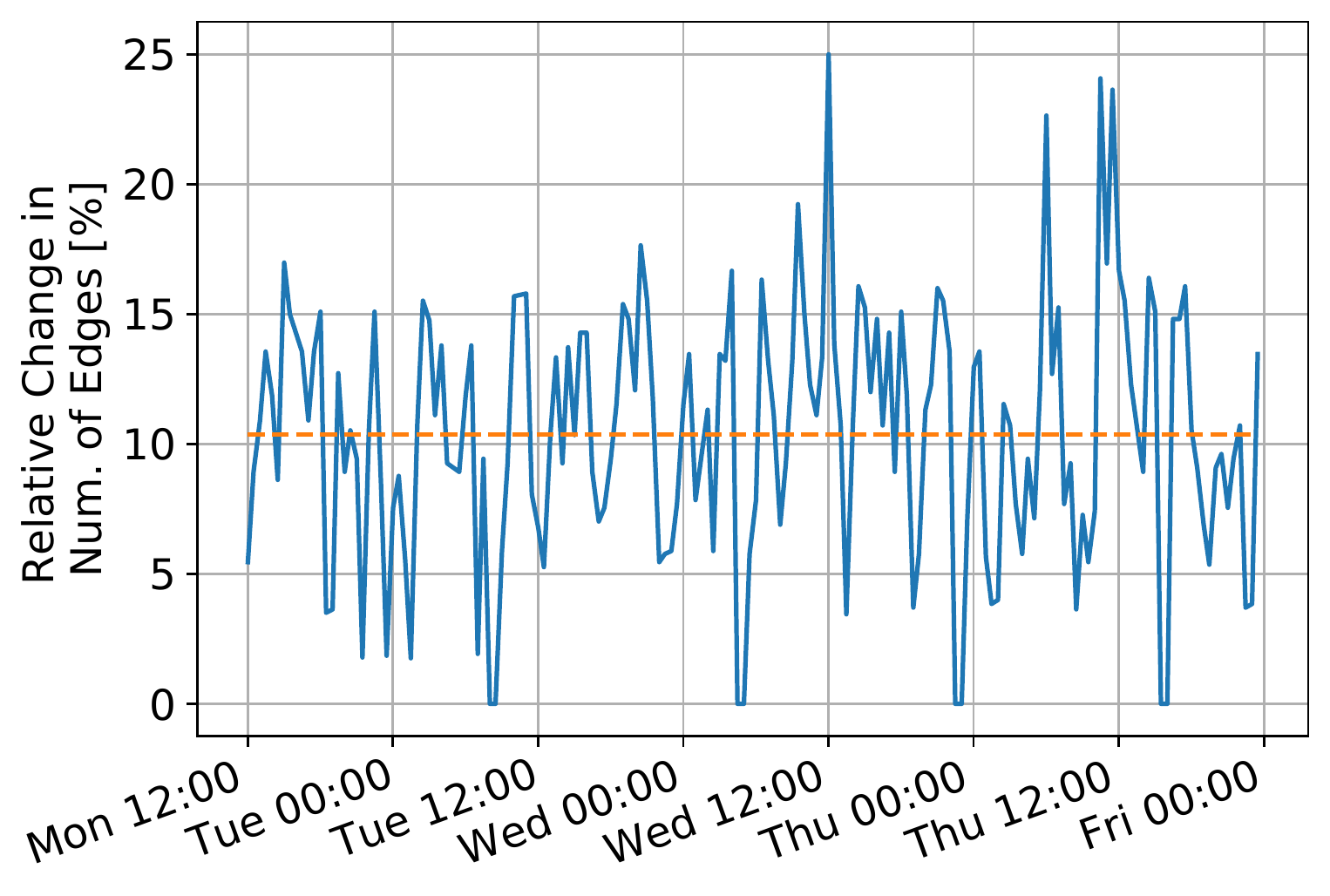}
         \caption{\capt{Link connectivity change over time}.}
         \label{subfig:pi-edge-change}
     \end{subfigure}
    \caption{
        \capt{We implement \system to compute schedules for a real IoT network}. The network exhibits a high rate of change in its connectivity between two subsequent link collection time periods. The orange dashed line in \ref{subfig:pi-edge-change} shows an average of $10.4\%$.}
    \label{fig:pi-setup}
\end{figure}

\fakepar{Setup} We use an indoor \ac{IoT} testbed consisting of 24 Zolertia Firefly devices (see Figure~\ref{subfig:pi-testbed-topo}). The devices run the Contiki-NG operating system~\cite{oikonomou22contiking}, communicate using IPv6 over IEEE 802.15.4 \ac{TSCH}~\cite{duquennoy17tsch}, and use RPL as routing protocol~\cite{winter2012rpl}. We collect the link connectivity among the \ac{IoT} nodes every 30 min over a period of four days. We assume there is a link between any given pair of nodes if there is a signal strength of at least \SI{-75}dBm for carrier provisioning. The \ac{IoT} network exhibits a dynamic change of node connectivity over the observed period of time (see Figure~\ref{subfig:pi-edge-change}). We augment each of the collected network topologies with randomly assigned tags in the range $T\in\{10, 25, 50, 75, 85\}$, $100$ assignments per $T$ value. 
 
\fakepar{Metrics} Apart from the \emph{carriers saved} metric employed in Sec.~\ref{subsec:test-set-perf} and \ref{subsec:generalization-perf}, we also include the average energy per tag interrogation $\Tilde{E}$, which is the total energy for tag interrogations $E_{tot}$ divided by the number of tags in the network. Based on Figure~\ref{subfig:timeslot-spec}, $\Tilde{E}$ is given by:  
\begin{equation}
    \Tilde{E} = \frac{E_{tot}}{T} = P_{tx}t_{tx}\!+\!P_{rx}\left(\frac{C}{T}t_{req}\!+\!t_{rx}\right)\!+\!P_{tx}\left(t_{req}\!+\!\frac{C}{T}t_{cg}\right) \text{,}
    \label{eq:avg-energy-pertag}
\end{equation}
where $C$ is the number of carriers used in the schedule, $T$ is the number of tags in the network, and both $P_{rx}$ and $P_{rx}$ correspond to the radio power at transmit and receive mode, respectively. We adopt $P_{rx}=72mW$, $P_{tx}=102mW$ based on the Firefly's reference values. Moreover, we assume $t_{req}=t_{tx}=128\mu s$, $t_{rx}=256\mu s$, and $t_{cg}=15.75 ms$~\cite{PerezPenichet2020afast}. 
 
\begin{figure}
     \centering
     \begin{subfigure}[b]{0.48\linewidth}
         \centering
         \includegraphics[width=0.98\linewidth]{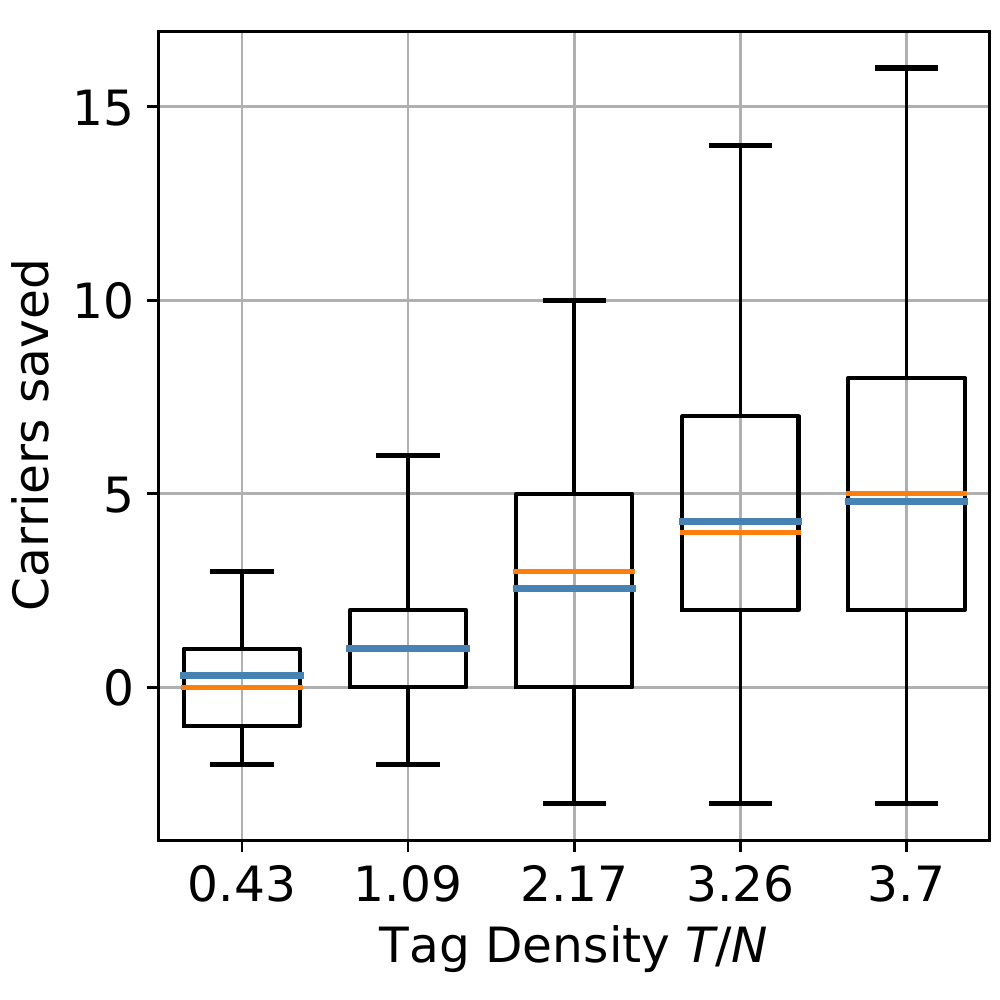}
         \caption{\capt{Carriers saved compared to TagAlong heuristic.}}
         \label{subfig:pi-testbed-carr-saved}
     \end{subfigure}
     \hfill
     \begin{subfigure}[b]{0.45\linewidth}
         \centering
         \includegraphics[width=0.98\linewidth]{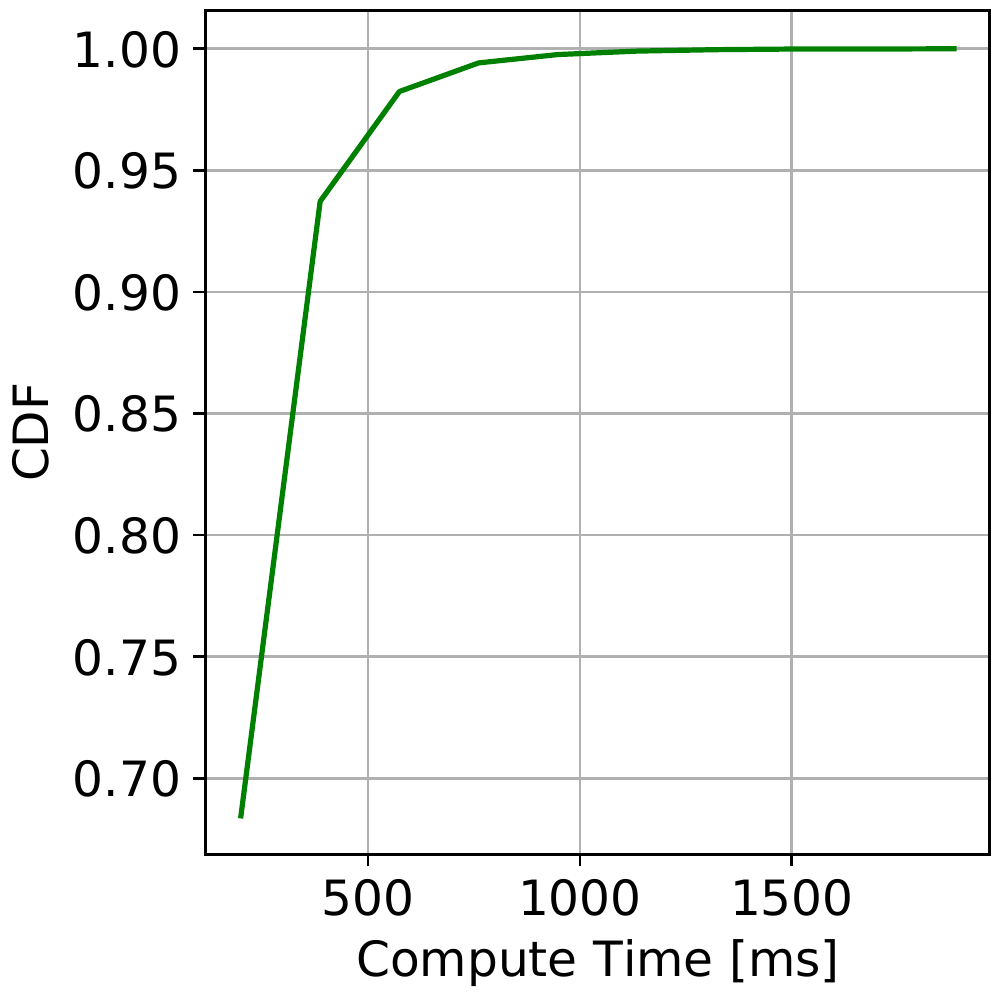}
         \caption{\capt{\system's runtime across all tag densities.}}
         \label{subfig:pi-runtime}
     \end{subfigure}
     \vfill
     \begin{subfigure}[b]{0.96\linewidth}
         \centering
         \includegraphics[width=\linewidth]{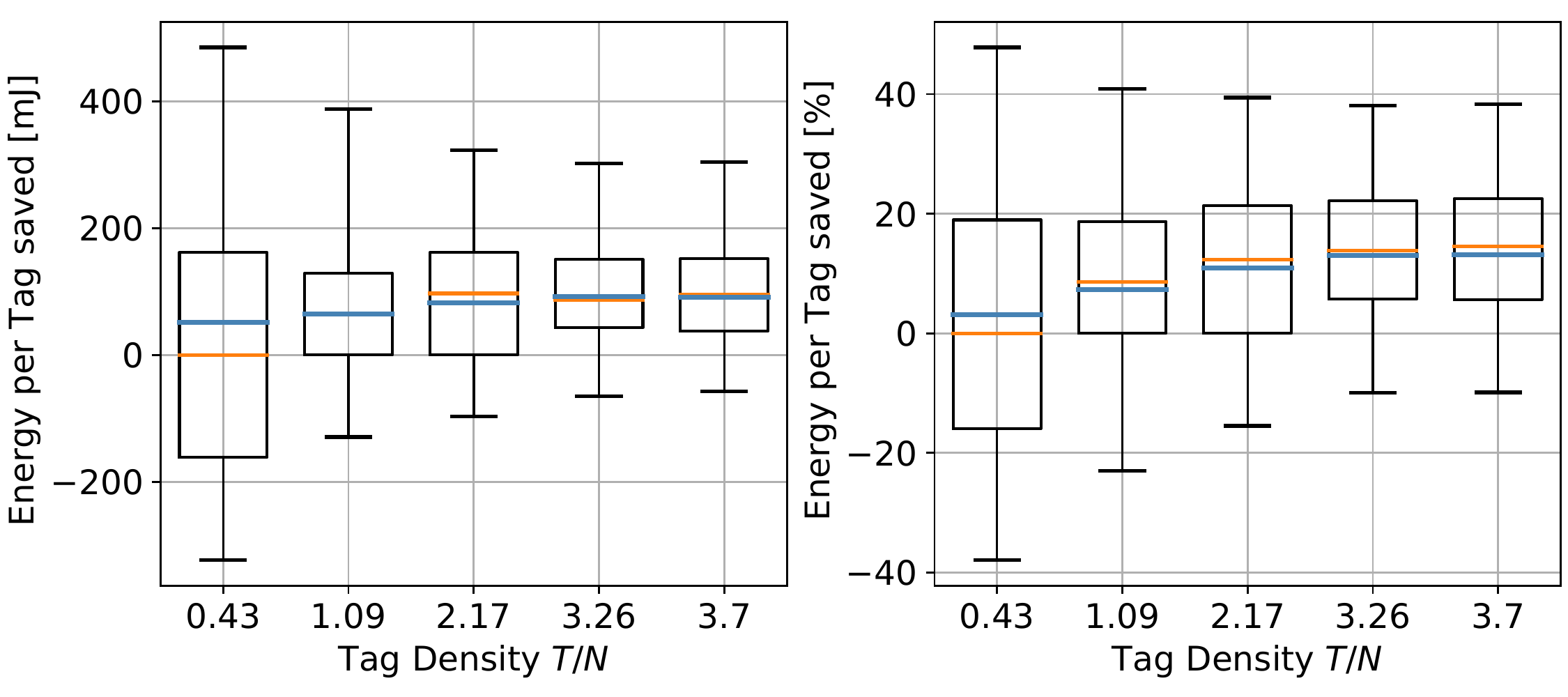}
         \caption{\capt{\system achieves $13\%$ in average and up to $51\%$ reduction of per-tag energy consumption compared to TagAlong for high tag densities}. }
         \label{subfig:pi-testbed-energy-saved}
     \end{subfigure}
    \caption{
        \capt{\system achieves high energy savings compared to TagAlong, even for high tag densities.} We successfully deployed \system to compute interrogation schedules for a real \ac{IoT} network of $N\!=\!24$ nodes and varying number of tags $T$. The blue and the orange line represent the mean and the median, respectively. Box extents delimit 25 and 75 percentiles. Whiskers delimit 5 and 99 percentiles.
        }
    \label{fig:pi-testbed-all}
    \vspace*{-0.5cm}
\end{figure}

\fakepar{Results} Figure~\ref{fig:pi-testbed-all} summarizes the results of deploying \system to compute schedules on the real \ac{IoT} network topologies compared to the TagAlong scheduler for different tag densities $\frac{T}{N}$. \system shows a similar behaviour of \emph{carriers saved} in Figure~\ref{subfig:pi-testbed-carr-saved} to those seen in Figure~\ref{fig:scalability-all}. Moreover, our scheduler achieves average savings in $\Tilde{E}$ compared to TagAlong above $10.96\%$ for $\frac{T}{N}\geq 2.17$, with up to $51.64\%$ maximum energy savings. Finally, Figure~\ref{subfig:pi-runtime} exhibits runtimes similar to those observed in Figure~\ref{fig:runtime}.  

\section{Related Work \& Discussion}\label{sec:related_work}

Our work is relevant both for 
scheduling in backscatter networks and
for supervised \ac{ML} applied to communications; in particular, to problems of a
combinatorial nature.

Many recent related efforts advance backscatter communications and battery-free
networks~\cite{kellogg_passive_2016,talla2017lora,iyer_inter-technology_2016,ensworth_every_2015,kellogg2014wi,zhang_freerider_2017,jansen_multihopbackscatter_2019,nikitin_tagtotag_2012,karimi_design_2017,geissdoerfer_bootstrapping_2021},
but few of these address the efficient provision of unmodulated carriers.
Pérez-Penichet~et~al. demonstrate TagAlong, a complete system with a
polynomial-time heuristic to compute interrogation schedules for backscatter
devices~\cite{perez-penichet_tagalong_2020,PerezPenichet2020afast}. Like our
work, TagAlong exploits knowledge of the structural properties of the wireless
network for fast scheduling. However, TagAlong's carefully designed algorithm
produces wasteful suboptimal schedules.
Van~Huynh~et~al.~\cite{huynh_optimalTS_2018} employ numerical analysis to
optimize \ac{RF} energy harvesting tags. By contrast, our work focuses on
communication aspects and remains independent of the energy harvesting modality.
Carrier scheduling resembles the Reader Collision Problem in RFID
systems~\cite{liu_season_2011,hamouda_reader_2011,chen_time-efficient_2012} in
that both need to avoid carrier collisions. These works focus on the monostatic
backscatter configuration (co-located carrier generator and receiver), whereas our work focuses on the bi-static configuration (separated carrier
generators and receivers).  The bi-static setting leads to a different
optimization problem and our focus is on resource optimization rather than mere
collision avoidance. 

Previous efforts in communications employ reinforcement learning with \acp{GNN}
to solve combinatorial scheduling problems, mostly on fixed-size networks or static
environments~\cite{Wang2019crowdcast, Zhang2019ReLes, Bhattacharyya2019qflow}.
By contrast, our work focuses on a single solution tackling variable-size inputs
and outputs, adequate for a multitude of varying conditions.
Also novel in our work is that we employ supervised \ac{ML} to solve the
\ac{COP}; to the best of our knowledge, this is a new approach within backscatter
communications.
Yet another novelty in our work is our strategy of restricting 
the solution space of the \ac{COP} to boost 
the trained model's performance and scalability properties.

\ac{ML} methods have been applied to \acp{COP} over graphs in the past
years~\cite{vesselinova2020learning}, for
both reinforcement~\cite{Dai2017learningcops, Manchanda2020learning} and
supervised learning~\cite{Vinyals2015pointernets, Li2018copsgcn}. 
Similar to our work, Vinyals~et~al.~\cite{Vinyals2015pointernets} 
implement an attention-based sequence-to-sequence model that learns from optimal solutions to solve the traveling salesperson problem. Likewise, Li~et~al.~\cite{Li2018copsgcn} employ 
\acp{GNN}~\cite{Kipf2017gcn, Defferrard2016gcnspectral} 
to solve three traditional \ac{COP}s
with a supervised approach. 


Our symmetry-breaking approach introduces bias in the scheduler: 
nodes with lower IDs would deplete their batteries faster, given that they will be selected first as carrier generators.
Far from a drawback, we believe this can be exploited as a feature at the application level. For instance, one could load-balance carrier scheduling over time, or schedule mains-powered carrier generators whenever possible instead of battery-powered ones simply by ordering the IDs in descending order of priority.

Finally, we believe that our approach to deal with multiple solutions in the scheduling
problem could have far-reaching implications in solving the broad class of
graph-related NP-hard \acp{COP} (such as traveling salesperson) using
supervised \ac{ML} techniques.

\section{Conclusion}\label{sec:conclusion}

\system is a scheduler that employs \acp{GNN} to
schedule a network of \ac{IoT} devices interoperating with battery-free
backscatter tags.  \system leverages self-attention \acp{GNN} to overcome the
challenges posed by the graph representing the problem and by
variable-size inputs and outputs.  Our symmetry-breaking strategy succeeds in
training \system to mimic the behavior of an optimal scheduler.
\system exhibits strong generalization capabilities to problem instances up to
six times larger than those used in training and can compute schedules,
requiring on average $\SIrange[range-phrase = -]{7}{10}{\percent}$ and up to $50\%$ fewer carriers than an existing,
carefully crafted heuristic, even for the largest problem instances considered.
More importantly, our scheduler performs within 3\% of the optimal on the
average number of carrier slots but with polynomial time complexity; lowering computation times from hours to
fractions of a second.  Our work advances the development of
practical and more efficient backscatter networks. This, in turn, paves the way for wider employment in a large range of environments that today pose problems of great difficulty and importance.


\bibliographystyle{./bibliography/ACM-Reference-Format}
\bibliography{./bibliography/refs}


\end{document}